\documentclass[runningheads]{llncs}
\usepackage[latin9]{inputenc}
\usepackage[paperwidth=146mm,paperheight=217mm]{geometry}
\usepackage{array}
\usepackage{float}
\usepackage{textcomp}
\usepackage{multirow}
\usepackage{amsmath}
\usepackage{amssymb}
\usepackage{graphicx}
\usepackage{wasysym}
\usepackage{algpseudocode}

\makeatletter

\providecommand{\tabularnewline}{\\}
\floatstyle{ruled}
\newfloat{algorithm}{tbp}{loa}
\providecommand{\algorithmname}{Algorithm}
\floatname{algorithm}{\protect\algorithmname}

\usepackage{comment}
\usepackage{color}
\usepackage{rotating}

\usepackage{tikz}

\newcommand\blfootnote[1]{%
	\begingroup
	\renewcommand\thefootnote{}\footnote{#1}%
	\addtocounter{footnote}{-1}%
	\endgroup
}

\mainmatter 
\def\ECCVSubNumber{3977}  
\titlerunning{AABO}
\authorrunning{Wenshuo Ma et al.}  
\author{Wenshuo Ma\inst{1}, Tingzhong Tian\inst{1}, Hang Xu\inst{\dagger}\inst{2},\\ Yimin Huang\inst{2}, Zhenguo Li\inst{2}} 
\institute{Tsinghua University \and Huawei Noah's Ark Lab}

\makeatother

\begin{document}
\title{AABO: Adaptive Anchor Box Optimization for Object Detection via Bayesian
Sub-sampling}
\maketitle
\begin{abstract}
\blfootnote{$\dagger$ Corresponding Author: xbjxh@live.com}
Most state-of-the-art object detection systems follow an anchor-based
diagram. Anchor boxes are densely proposed over the images and the
network is trained to predict the boxes position offset as well as
the classification confidence. Existing systems pre-define anchor
box shapes and sizes and ad-hoc heuristic adjustments are used to
define the anchor configurations. However, this might be sub-optimal
or even wrong when a new dataset or a new model is adopted. In this
paper, we study the problem of automatically optimizing anchor boxes
for object detection. We first demonstrate that the number of anchors,
anchor scales and ratios are crucial factors for a reliable object
detection system. By carefully analyzing the existing bounding box patterns
on the feature hierarchy, we design a flexible and tight hyper-parameter
space for anchor configurations. Then we propose a novel hyper-parameter
optimization method named AABO to determine more appropriate anchor
boxes for a certain dataset, in which Bayesian Optimization and sub-sampling
method are combined to achieve precise and efficient anchor configuration
optimization. Experiments demonstrate the effectiveness of our proposed
method on different detectors and datasets, e.g. achieving around
2.4\% mAP improvement on COCO, 1.6\% on ADE and 1.5\% on VG, and the
optimal anchors can bring 1.4\% $\sim$ 2.4\% mAP improvement
on SOTA detectors by only optimizing anchor configurations, e.g. boosting
Mask RCNN from 40.3\% to 42.3\%, and HTC detector from 46.8\%
to 48.2\%. \keywords{Object detection, Hyper-parameter optimization,
Bayesian optimization, Sub-sampling}
\end{abstract}

\section{Introduction}

Object detection is a fundamental and core problem in many computer
vision tasks and is widely applied on autonomous vehicles \cite{chabot2017deep},
surveillance camera \cite{luo2014switchable}, facial recognition
\cite{bhagavatula2017faster}, to name a few. Object detection aims
to recognize the location of objects and predict the associated class
labels in an image. Recently, significant progress has been made on
object detection tasks using deep convolution neural network \cite{liu2016ssd,redmon2016you,ren2015faster,lin2017feature}.
In many of those deep learning based detection techniques, anchor
boxes (or default boxes) are the fundamental components, serving as
initial suggestions of object's bounding boxes. Specifically, a large
set of densely distributed anchors with pre-defined scales and aspect
ratios are sampled uniformly over the feature maps, then both shape
offsets and position offsets relative to the anchors, as well as classification
confidence, are predicted using a neural network.

While anchor configurations are rather critical hyper-parameters of
the neural network, the design of anchors always follows straight-forward
strategies like handcrafting or using statistical methods such as
clustering. Taking some widely used detection frameworks for instance,
Faster R-CNN \cite{ren2015faster} uses pre-defined anchor shapes
with 3 scales ($128^{2},256^{2},512^{2}$) and 3 aspect ratios ($1:1$, $1:2$, $2:1$), and YOLOv2 \cite{redmon2017yolo9000} models anchor
shapes by performing k-means clustering on the ground-truth of bounding
boxes. And when the detectors are extended to a new certain problem,
anchor configurations must be manually modified to adapt the property
and distribution of this new domain, which is difficult and inefficient,
and could be sub-optimal for the detectors.

While it is irrational to determine hyper-parameters manually, recent
years have seen great development in hyper-parameter optimization
(HPO) problems and a great quantity of HPO methods are proposed. The
most efficient methods include Bayesian Optimization (BO) and bandit-based
policies. BO iterates over the following three steps: a) Select the
point that maximizes the acquisition function. b) Evaluate the objective
function. c) Add the new observation to the data and refit the model,
which provides an efficacious method to select promising hyper-parameters
with sufficient resources. Different from BO, Bandit-based policies
are proposed to efficiently measure the performance of hyper-parameters.
Among them, Hyperband \cite{Li2016Hyperband} (HB) makes use of cheap-to-evaluate
approximations of the acquisition function on smaller budgets, which
calls SuccessiveHalving \cite{jamieson2016non} as inner loop to identify
the best out of $n$ randomly-sampled configurations. Bayesian Optimization
and Hyperband (BOHB) introduced in \cite{Falkner2018} combined these
two methods to deal with HPO problems in a huge search space, and
it is regarded as a very advanced HPO method. However, BOHB is less
applicable to our anchor optimization problems, because the appropriate
anchors for small objects are always hard to converge, then the optimal
anchor configurations could be early-stopped and discarded by SuccesiveHalving.

In this paper, we propose an adaptive anchor box optimization method
named AABO to automatically discover optimal anchor configurations,
which can fully exploit the potential of the modern object detectors.
Specifically, we illustrate that anchor configurations such as the
number of anchors, anchor scales and aspect ratios are crucial factors
for a reliable object detector, and demonstrate that appropriate anchor
boxes can improve the performance of object detection systems. Then
we prove that anchor shapes and distributions vary distinctly across
different feature maps, so it is irrational to share identical anchor
settings through all those feature maps. So we design a tight and
adaptive search space for feature map pyramids after meticulous analysis
of the distribution and pattern of the bounding boxes in existing
datasets, to make full use of the search resources. After optimizing
the anchor search space, we propose a novel hyper-parameter optimization
method combining the benefits of both Bayesian Optimization and sub-sampling
method. Compared with existing HPO methods, our proposed approach
uses sub-sampling method to estimate acquisition function as accurately
as possible, and gives opportunity to the configuration to be assigned
with more budgets if it has chance to be the best configuration, which
can ensure that the promising configurations will not be discarded
too early. So our method can efficiently determine more appropriate
anchor boxes for a certain dataset using limited computation resources,
and achieves better performance than previous HPO methods such as
random search and BOHB.

We conduct extensive experiments to demonstrate the effectiveness
of our proposed approach. Significant improvements over the default
anchor configurations are observed on multiple benchmarks. In particular,
AABO achieves 2.4\% mAP improvement on COCO, 1.6\% on ADE and 1.5\%
on VG by only changing the anchor configurations, and consistently
improves the performance of SOTA detectors by 1.4\% $\sim$ 2.4\%, e.g. boosts Mask RCNN \cite{He2017Mask} from 40.3\% to 42.3\%
and HTC \cite{chen2019hybrid} from 46.8\% to 48.2\% in terms of mAP.

\section{Related Work}

\subsubsection{Anchor-Based Object Detection.}

Modern object detection pipelines based on CNN can be categorized
into two groups: One-stage methods such as SSD \cite{liu2016ssd}
and YOLOv2 \cite{redmon2017yolo9000}, and two-stage methods such
as Faster R-CNN \cite{ren2015faster} and R-FCN \cite{dai2016r}.
Most of those methods make use of a great deal of densely distributed
anchor boxes. In brief, those modern detectors regard anchor boxes
as initial references to the bounding boxes of objects in an image.
The anchor shapes in those methods are typically determined by manual
selection \cite{liu2016ssd,ren2015faster,dai2016r} or naive clustering
methods \cite{redmon2017yolo9000}. Different from the traditional
methods, there are several works focusing on utilizing anchors more
effectively and efficiently \cite{Tong2018MetaAnchor,Zhong2018}.
MetaAnchor \cite{Tong2018MetaAnchor} introduces meta-learning to
anchor generation, which models anchors using an extra neural network
and computes anchors from customized priors. However, the network
becomes more complicated. Zhong et al. \cite{Zhong2018} tries to learn
the anchor shapes during training via a gradient-based method while
the continuous relaxation may be not appropriate.

\subsubsection{Hyper-paramter Optimization.}

\noindent Although deep learning has achieved great successes in a
wide range, the performance of deep learning models depends strongly
on the correct setting of many internal hyper-parameters, which calls
for an effective and practical solution to the hyper-parameter optimization
(HPO) problems. Bayesian Optimization (BO) has been successfully applied
to many HPO works. For example, \cite{Snoek2012Practical} obtained
state-of-the-art performance on CIFAR-10 using BO to search out the
optimal hyper-parameters for convolution neural networks. And \cite{mendoza2016towards}
won 3 datasets in the 2016 AutoML challenge by automatically finding
the proper architecture and hyper-parameters via BO methods. While
BO approach can converge to the best configurations theoretically,
it requires an awful lot of resources and is typically computational
expensive. Compared to Bayesian method, there exist bandit-based configuration
evaluation approaches based on random search such as Hyperband \cite{Li2016Hyperband},
which could dynamically allocate resources and use SuccessiveHalving
\cite{jamieson2016non} to stop poorly performing configurations.
Recently, some works combining Bayesian Optimization with Hyperband
are proposed like BOHB \cite{Falkner2018}, which can obtain strong
performance as well as fast convergence to optimal configurations.
Other non-parametric methods that have been proposed include $\varepsilon$-greedy
and Boltzmann exploration \cite{sutton2018reinforcement}. \cite{Chan2019}
proposed an efficient non-parametric solution and proved optimal efficiency
of the policy which would be extended in our work. However, there
exist some problems in those advanced HPO methods such as expensive
computation in BO and early-stop in BOHB.

\section{The Proposed Approach}

\subsection{Preliminary Analysis}

As mentioned before, mainstream detectors, including one-stage and
two-stage detectors, rely on anchor boxes to provide initial guess
of the object's bounding box. And most detectors pre-define anchors
and manually modify them when applied on new datasets. We believe
that those manual methods can hardly find optimal anchor configurations
and sub-optimal anchors will prevent the detectors from obtaining
the optimal performance. To confirm this assumption, we construct
two preliminary experiments.

\begin{figure}[tb]
\begin{centering}
\par\end{centering}
\begin{centering}
\includegraphics[scale=0.25]{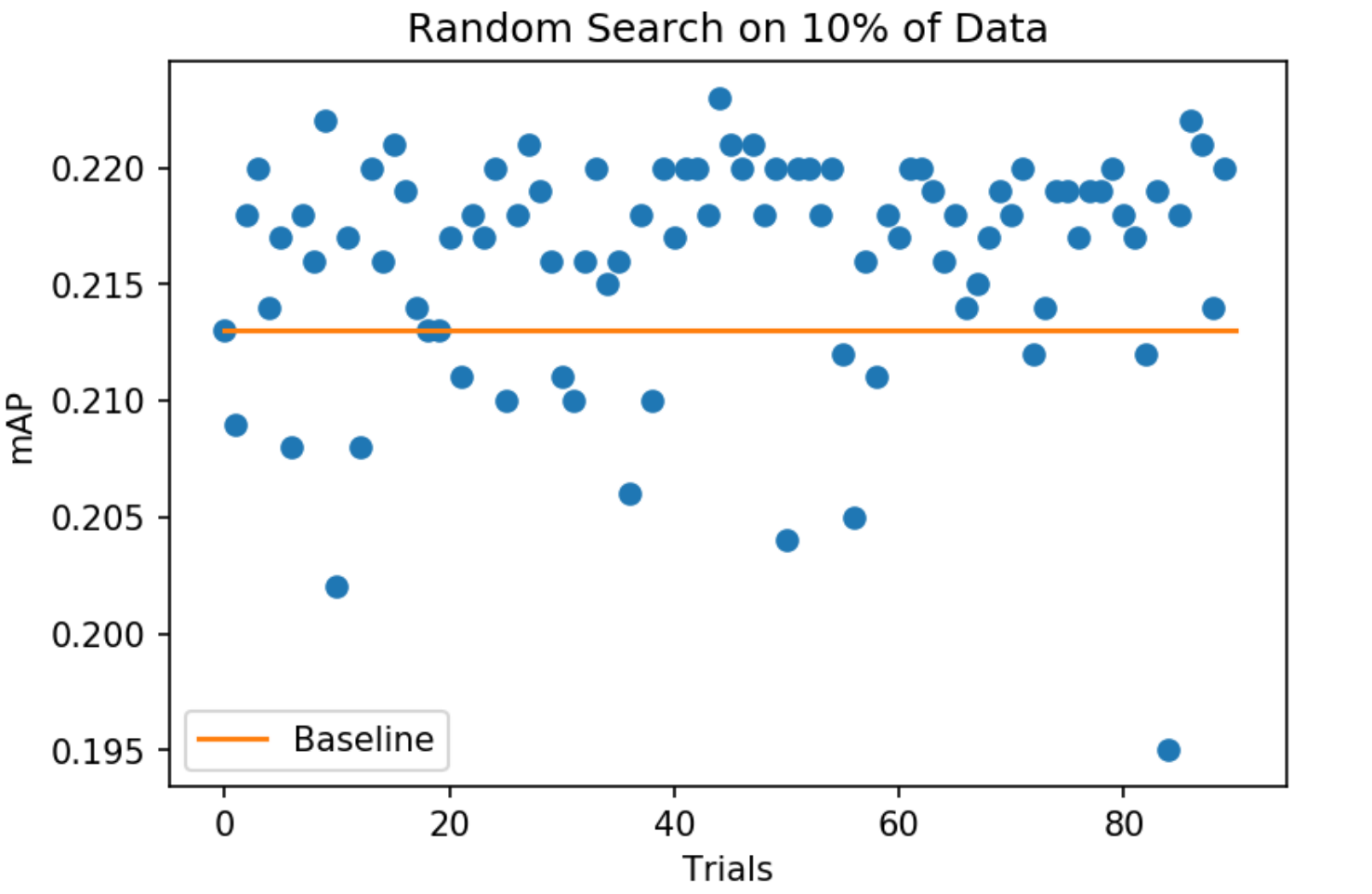}\includegraphics[scale=0.25]{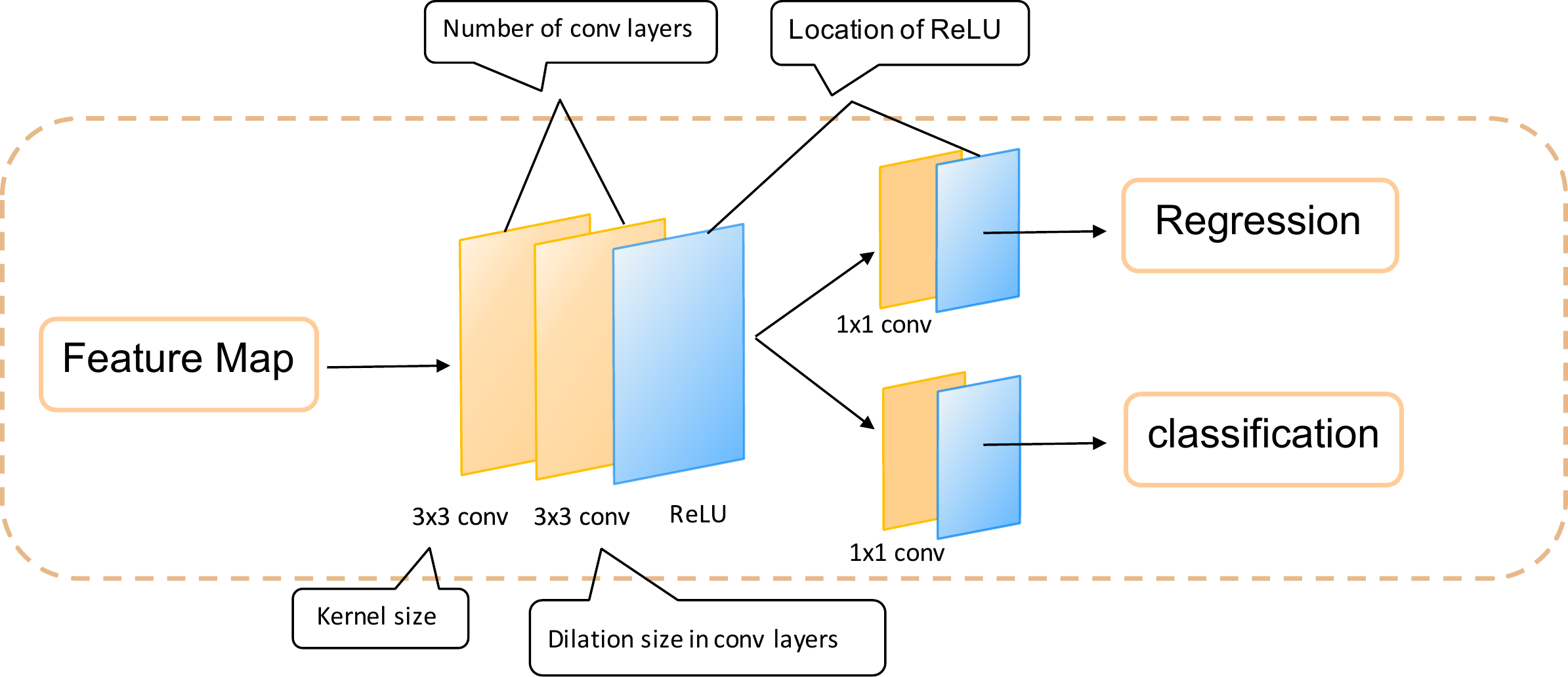}
\par\end{centering}

\caption{\label{fig:pretest-1}(Left) The performance of Faster-R-CNN \cite{ren2015faster}
under different anchor configurations on 10\% data of COCO \cite{lin2014microsoft}.
Randomly-sampled anchors significantly influence the performance of
the detector. (Right) The search space for RPN Head in FPN \cite{lin2017feature}
consists of the number of convolution layers, kernel size, dilation size,
and the location of nonlinear activation functions ReLU, which are
illustrated in the dialogue boxes.}

\centering{}
\end{figure}

\textbf{Default Anchors Are Not Optimal.} We randomly sample 100 sets of different anchor settings, each with
3 scales and 3 ratios. Then we examine the performance of Faster-RCNN
\cite{ren2015faster} under those anchor configurations. The results
are shown in Figure \ref{fig:pretest-1}.

It's obvious that compared with default anchor setting (3 anchor scales:
$128^{2},256^{2},512^{2}$ and 3 aspect ratios: $1:1$, $1:2$, and
$2:1$), randomly-sampled anchor settings could significantly influence
the performance of the detector, which clearly demonstrates that the
default anchor settings may be less appropriate and the optimization
of anchor boxes is necessary.

\textbf{Anchors Influence More Than RPN Structure.} Feature Pyramid Networks \cite{lin2017feature} (FPN) introduces a
top-down pathway and lateral connections to enhance the semantic representation
of low-level features and is a widely used feature fusion scheme in
modern detectors. In this section, we use BOHB \cite{Falkner2018}
to search RPN head architecture in FPN as well as anchor settings
simultaneously. The search space of RPN Head is illustrated in Figure
\ref{fig:pretest-1}. The searched configurations, including RPN head
architecture and anchor settings, are reported in the appendix. Then
we analyze the respective contributions of RPN head architecture and
anchor settings, and the results are shown in Table \ref{tab:rpn-anchor-con}.
Here, mean Average Precision is used to measure the performance and
is denoted by mAP.

\begin{table}[tb]
{\footnotesize{}\caption{\label{tab:rpn-anchor-con}The respective contributions of RPN head
architecture and anchor configurations. Anchor optimization could
obviously improve the mAP of the detectors while architecture optimization
produces very little positive effect, or even negative effect. All
the experiments are conducted on COCO, using FPN as detector. Note
that the search space for RPN head architecture and anchor settings
are both relatively small, then the performance improvements are not
that significant.}
}{\footnotesize\par}
\begin{centering}
\par\end{centering}
\begin{centering}
\tabcolsep 0.05in{\footnotesize{}}%
\begin{tabular}{c|cccccc}
\hline 
{\scriptsize{}Performance} & {\scriptsize{}mAP} & {\scriptsize{}AP$_{50}$} & {\scriptsize{}AP$_{75}$} & {\scriptsize{}AP$_{S}$} & {\scriptsize{}AP$_{M}$} & {\scriptsize{}AP$_{L}$}\tabularnewline
\hline 
{\scriptsize{}Baseline} & {\scriptsize{}36.4} & {\scriptsize{}58.2} & {\scriptsize{}39.1} & {\scriptsize{}21.3} & {\scriptsize{}40.1} & {\scriptsize{}46.5}\tabularnewline
{\scriptsize{}Only Anchor} & {\scriptsize{}37.1$^{+0.7}$} & {\scriptsize{}58.4} & {\scriptsize{}40.0} & {\scriptsize{}20.6} & {\scriptsize{}40.8} & {\scriptsize{}49.5}\tabularnewline
{\scriptsize{}Only Architecture} & {\scriptsize{}36.3$^{-0.1}$} & {\scriptsize{}58.3} & {\scriptsize{}38.9} & {\scriptsize{}21.6} & {\scriptsize{}40.2} & {\scriptsize{}46.1}\tabularnewline
{\scriptsize{}Anchor+Architecture} & \textbf{\scriptsize{}37.2}{\scriptsize{}$^{+0.8}$} & \textbf{\scriptsize{}58.7} & \textbf{\scriptsize{}40.2} & \textbf{\scriptsize{}20.9} & \textbf{\scriptsize{}41.3} & \textbf{\scriptsize{}49.1}\tabularnewline
\hline 
\end{tabular}{\footnotesize\par}
\par\end{centering}
\centering{}
\end{table}

The results in Table \ref{tab:rpn-anchor-con} illustrate that searching
for proper anchor configurations could bring more performance improvement
than searching for RPN head architecture to a certain extent.

Thus, the conclusion comes clearly that anchor settings affect detectors
substantially, and proper anchor settings could bring considerable
improvement than doing neural architecture search (NAS) on the architecture of RPN head. Those conclusions indicate that the optimization
of anchor configurations is essential and rewarding, which motivates
us to view anchor configurations as hyper-parameters and propose a
better HPO method for our anchor optimization case.

\subsection{Search Space Optimization for Anchors}

Since we have decided to search appropriate anchor configurations
to increase the performance of detectors over a certain dataset, one
critical problem is how to design the search space. In the preliminary
analysis, we construct two experiments whose search space is roughly
determined regardless of the distribution of bounding boxes. In this
section, we will design a much tighter search space by analyzing the
distribution characteristics of object's bounding boxes.

For a certain detection task, we find that anchor distribution satisfies
some properties and patterns as follows.

\textbf{Upper and Lower Limits of the Anchors.} Note that the anchor scale and anchor ratio are calculated from the
$width$ and $height$ of the anchor, which are not independent. Besides,
we discover that both anchor $width$ and $height$ are limited within
fixed values, denoted by $W$and $H$. Then the anchor ratio and scale
must satisfy constraints as follows:

{\small{}
\begin{equation}
\begin{cases}
\begin{array}{c}
scale=\sqrt{width*height}\\
ratio=height/width\\
width\leq W,\,height\leq H.
\end{array} & \text{}\end{cases}
\end{equation}
}{\small\par}

From the formulas above, we calculate the upper bound and lower bound
of the ratio for the anchor boxes:

{\small{}
\begin{equation}
\begin{array}{c}
\dfrac{scale^{2}}{W^{2}}\leq ratio\leq\dfrac{H^{2}}{scale^{2}}\end{array} .
\end{equation}
}{\small\par}

\begin{figure}[tb]
\begin{centering}
\par\end{centering}
\begin{centering}
\includegraphics[scale=0.195]{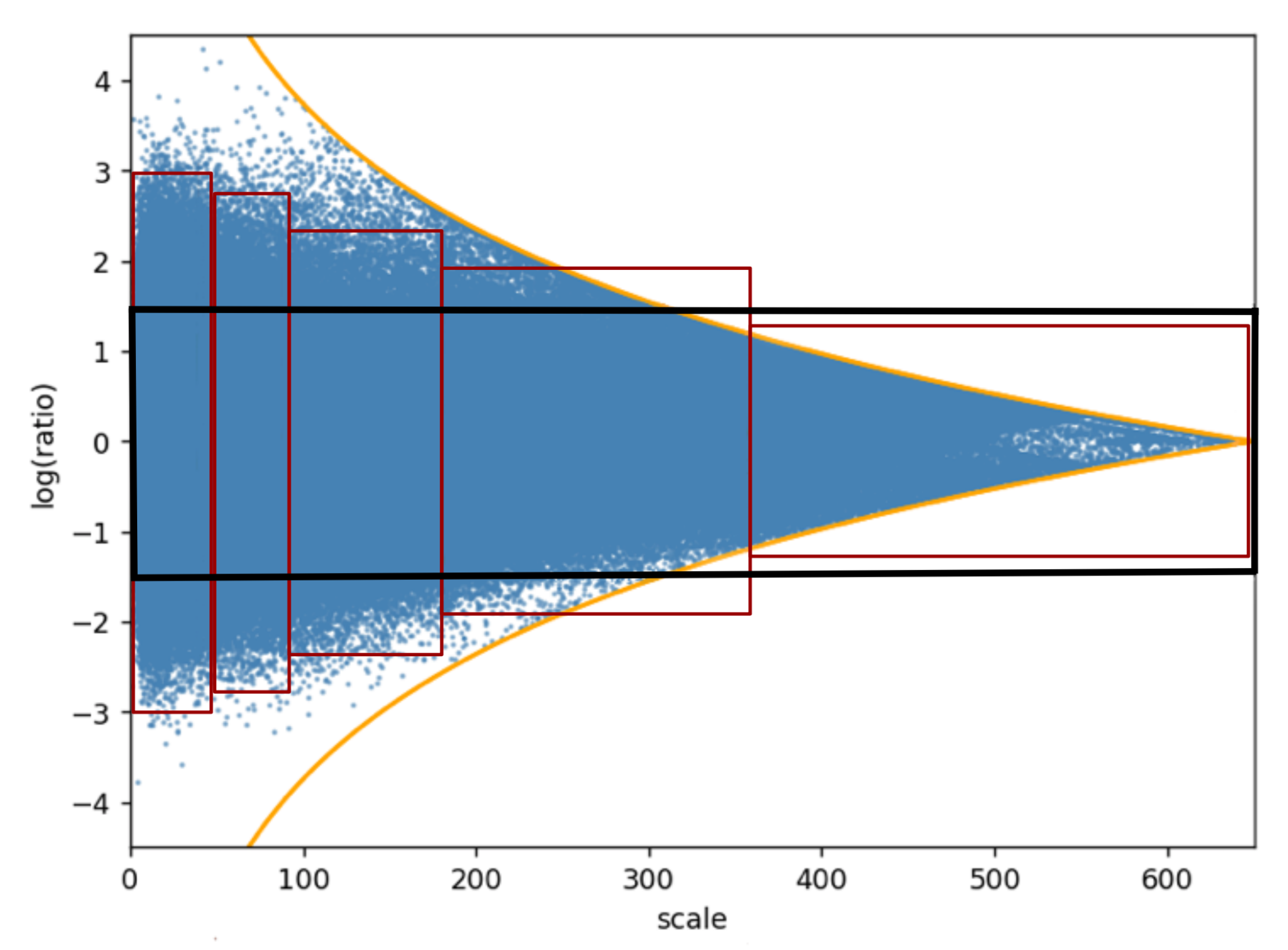}\includegraphics[scale=0.08]{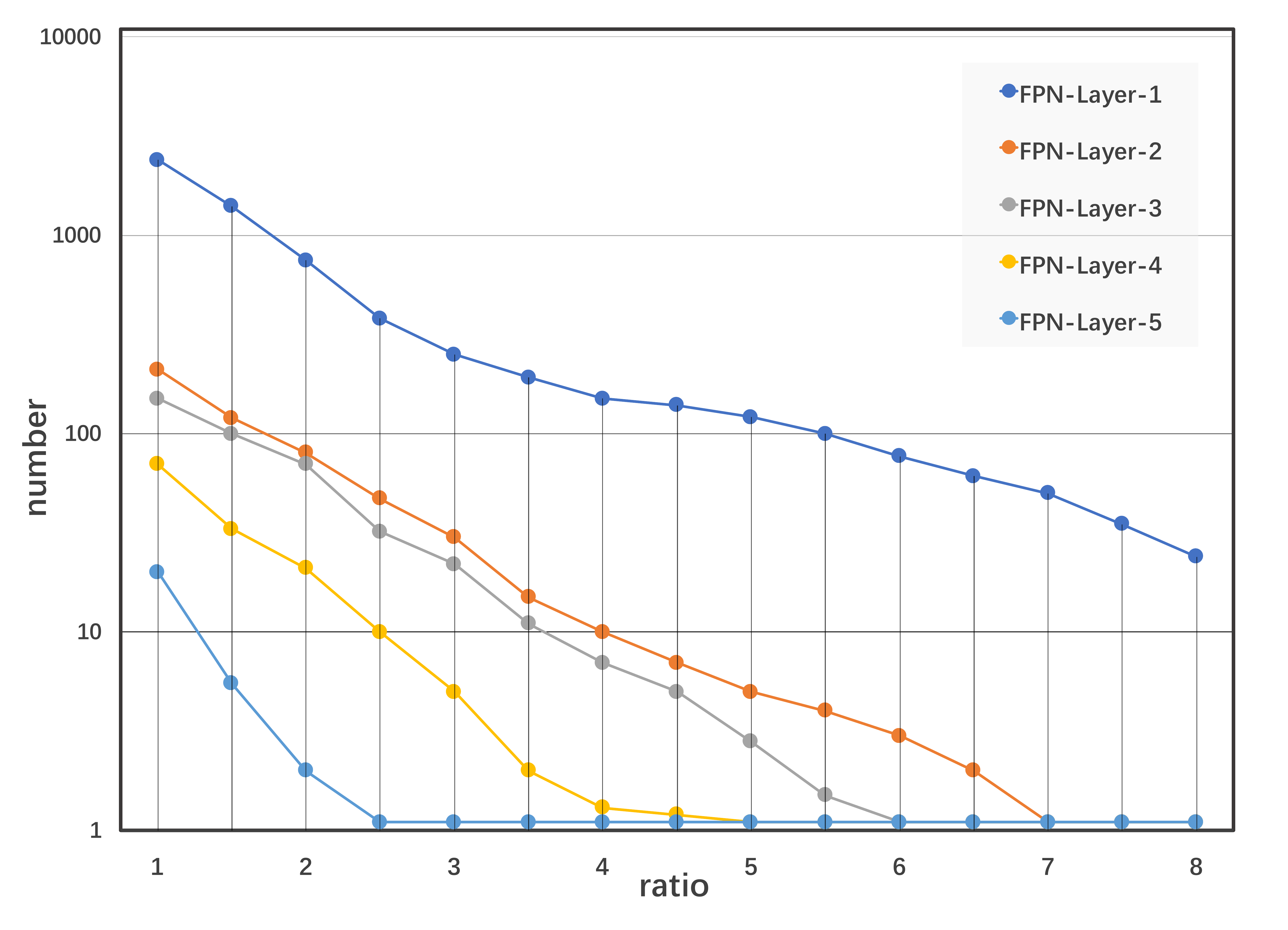}
\par\end{centering}
\begin{centering}
\caption{\label{fig:bound}(Left) Bounding boxes in COCO \cite{lin2014microsoft}
dataset apparently only distribute in a certain area determined by
the yellow curves. The region inside the black rectangle represents
the previous search space which is coarse and unreasonable as analyzed.
The intersecting regions between the 5 red rectangles and the anchor
distribution bounds are our designed feature-map-wise search space,
which is more accurate and adaptive. (Right) Numbers and shapes of
bounding boxes vary a lot across different feature maps. In this figure,
X-axis is the anchor ratio while Y-axis is the number of bounding
boxes. It's obvious that the number of bounding boxes decreases rapidly,
and the range of anchor ratios also becomes rather smaller in higher
feature map.}
\par\end{centering}
\centering{}
\end{figure}

Figure \ref{fig:bound} shows an instance of the distribution of bounding
boxes (the blue points) as well as the upper and lower bounds of anchors
(the yellow curves) in COCO \cite{lin2014microsoft} dataset. And
the region inside the black rectangle is the previous search space
using in preliminary experiments. We can observe that there exists
an area which is beyond the upper and lower bounds so that bounding
boxes won't appear, while the search algorithm will still sample anchors
here. So it's necessary to limit the search space within the upper
and lower bounds.

\textbf{Adaptive Feature-Map-Wised Search Space.} We then study the distribution of anchor boxes in different feature
maps of Feature Pyramid Networks (FPN)\cite{lin2017feature} and discover
that the numbers, scales and ratios of anchor boxes vary a lot across
different feature maps, shown in the right subgraph of Figure \ref{fig:bound}.
There are more and bigger bounding boxes in lower feature maps whose
receptive fields are smaller, less and tinier bounding boxes in higher
feature maps whose respective fields are wider.

As a result, we design an adaptive search space for FPN \cite{lin2017feature}
as shown in the left subgraph of Figure \ref{fig:bound}. The regions
within 5 red rectangles and the anchor distribution bounds represent
the search space for each feature map in FPN. As feature maps become
higher and smaller, the numbers of anchor boxes are less, as well
as the anchor scales and ratios are limited to a narrower range.

\begin{figure}[tb]
\begin{centering}
\par\end{centering}
\begin{centering}
\includegraphics[scale=0.165]{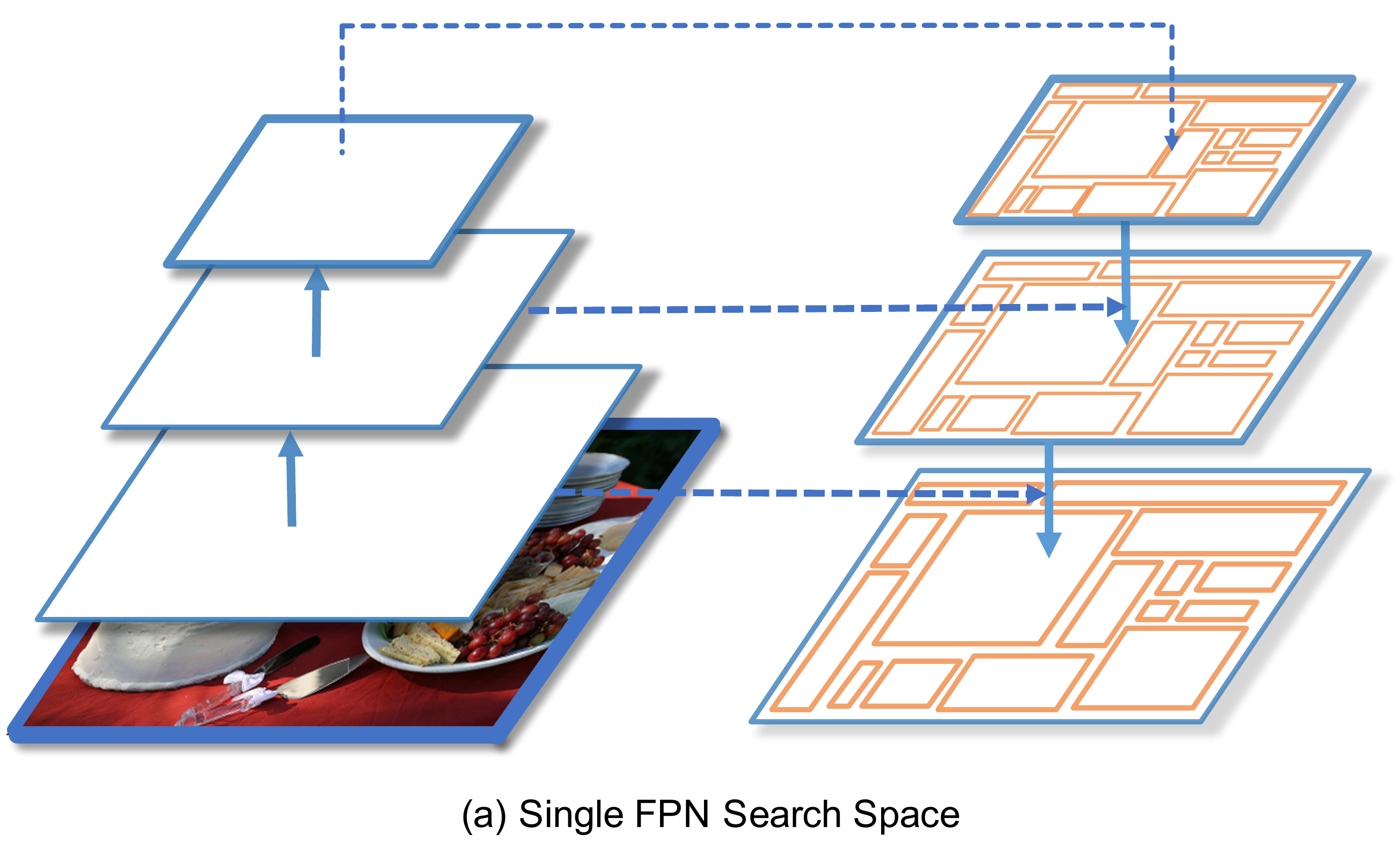}\includegraphics[scale=0.165]{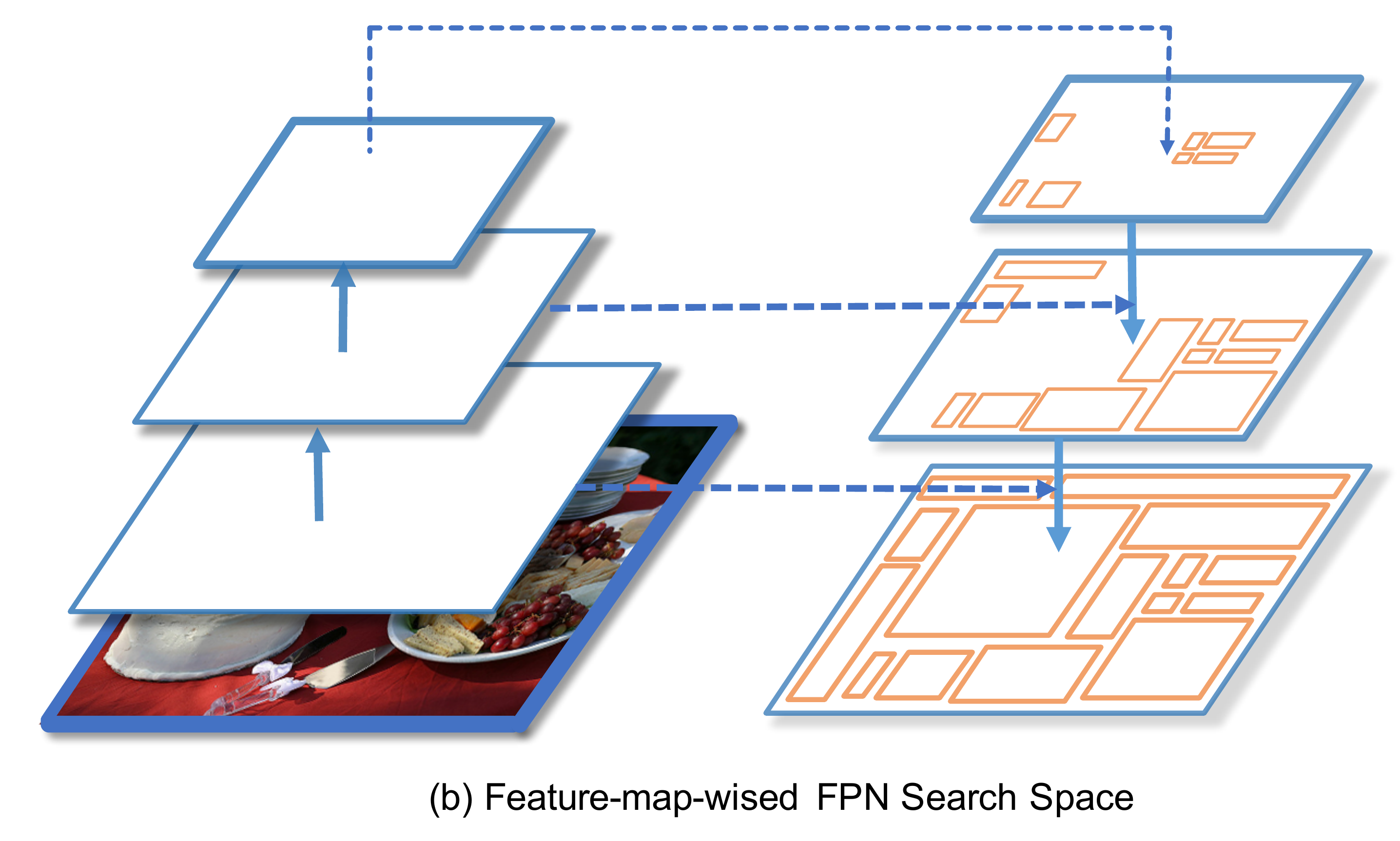}
\par\end{centering}
\begin{centering}
\par\end{centering}
\caption{\label{fig:new-feature maps} We design an adaptive feature-map-wised
search space for anchor configuration optimization. Compared to the
former single search space illustrated in (a), where there exist exactly
identical anchors among different feature maps, feature-map-wised
search space illustrated in (b) takes bounding-box distribution into
consideration, so extreme anchors are fewer in higher feature maps.
That is, in lower feature layers, there are more diverse, larger and
more anchors, while in higher feature layers, there are less diverse,
smaller and fewer anchors.}

\centering{}
\end{figure}

Compared with the initial naive search space, we define a tighter
and more adaptive search space for FPN \cite{lin2017feature}, as
shown in Figure \ref{fig:new-feature maps}. Actually, the new feature-map-wised
search space is much bigger than the previous one, which makes it
possible to select more flexible anchors and cover more diverse objects
with different sizes and shapes. Besides, the tightness of the search
space can help HPO algorithms concentrate limited resources on more
meaningful areas and avoid wasting resources in sparsely distributed
regions of anchors.

\subsection{Bayesian Anchor Optimization via Sub-sampling}

As described before, we regard anchor configurations as hyper-parameters
and try to use HPO method to choose optimal anchor settings automatically.
However, existing HPO methods are not suitable for solving our problems.
For random search or grid search, it's unlikely to find a good solution
because the search space is too big for those methods. For Hyperband or BOHB, the proper configurations for the small objects could
be discarded very early since the anchors of small objects are always
slowly-converged. So we propose a novel method which combines Bayesian
Optimization and sub-sampling method, to search out the optimal configurations
as quickly as possible.

\begin{figure}[t]
\begin{centering}
\par\end{centering}
\begin{centering}
\includegraphics[scale=0.26]{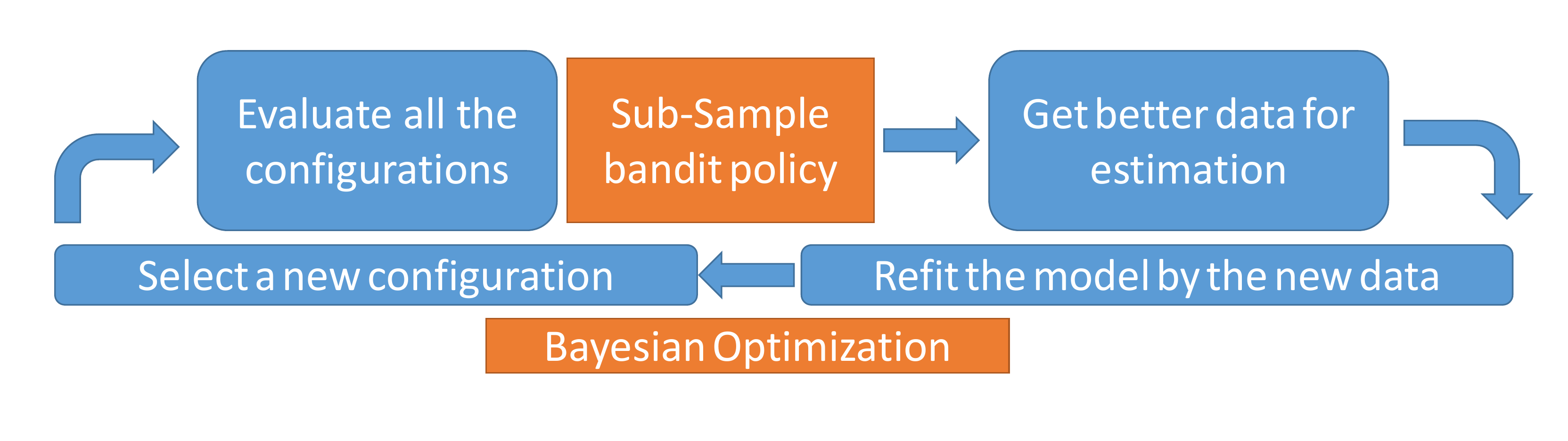}
\par\end{centering}
\begin{centering}
\caption{\label{fig:boss-1}Our proposed method iterates over the following
four steps: (a) Select the point that maximizes the acquisition function.
(b) Evaluate the objective function on the whole configurations with
Sub-Sampling policy. (c) Get more appropriate data for estimating
the densities in the model. (d) Add the new observation to the data
and refit the model.}
\par\end{centering}
\centering{}
\end{figure}

Specifically, our proposed approach makes use of BO to select potential
configurations, which estimates the acquisition function based on
the configurations already evaluated, and then maximizes the acquisition
fuction to identify promising new configurations. Meanwhile, sub-sampling
method is employed to determine which configurations should be allocated
more budgets, and explore more configurations in the search space.
Figure \ref{fig:boss-1} illustrates the process of our proposed method.
In conclusion, our approach can achieve good performance as well as
better speed, and take full advantage of models built on previous
budgets.

\textbf{Bayesian Optimization.} Bayesian Optimization (BO) is a sequential design strategy for parameter
optimization of black-box functions. In hyper-parameter optimization
problems, the validation performance of machine learning algorithms
is regarded as a function $f:\chi\rightarrow\mathbb{R}$ of hyper-parameters
$x\in\chi$, and hyper-parameter optimization problem aims to determine
the optimal $x_{*}\in argmin_{\chi}f(x)$ . In most machine learning
problems, $f(x)$ is unobservable, so Bayesian Optimization approach
treats it as a random function with a prior over it. Then some data
points are sampled and BO updates the prior and models the function
based on those gathering data points and evaluations. Then new data
points are selected and observed to refit the model function.

In our approach, we use Tree Parzen Estimator (TPE) \cite{Bergstra2011Algorithms}
which uses a kernel density estimator to model the probability density
functions $l(x)=p(y<\alpha|x,D)$ and $g(x)=p(y>\alpha|x,D)$ instead
of modeling function $p(f|D)$ directly, where $D=\{(x_{0},y_{0}),\ldots,(x_{n},y_{n})\}$
and $\alpha=\min\{y_{0},\ldots,y_{n}\}$, as seen in BOHB \cite{Falkner2018}.
Note that there exists a serious problem that the theory of Hyperband
only guarantees to return the best $x_{i}$ which has the smallest
$y_{i}$ among all these configurations, while the quality of other
configurations may be very poor. Consequently, larger responses lead
to an inaccurate estimation of $l(x)$, which plays an important role
in TPE. Thus, we need to propose a policy that has a better sequence
of responses $y_{i}$ to solve this problem.

\textbf{Sub-Sample Method.} To better explain the sub-sampling method used in our proposed approach,
we first introduce the standard multi-armed bandit problem with $K$
arms in this section. Recall the traditional setup for the classic
multi-armed bandit problem. Let $\mathcal{I}=\{1,2,\ldots,K\}$ be
a given set of $K\geq2$ arms. Consider a sequential procedure based
on past observations for selecting an arm to pull. Let $N_{k}$ be
the number of observations from the arm $k$, and $N=\sum_{k=1}^{K}N_{k}$
is the number of total observations. Observations $Y_{1}^{(k)},Y_{2}^{(k)},\ldots$,
$1\leq k\leq K$ are also called rewards from the arm $k$. In each
arm, rewards $\{Y_{t}^{(k)}\}_{t\geq1}$ are assumed to be independent
and identically distributed with expectation given by $\mathbb{E}(Y_{t}^{(k)})=\mu_{k}$
and $\mu_{*}=\max_{1\leq k\leq K}\mu_{k}$. For simplicity, assume
without loss of generality that the best arm is \emph{unique} which
is also assumed in \cite{perchet2013the} and \cite{Chan2019}.

A \emph{policy} $\pi=\{\pi_{t}\}$ is a sequence of random variables
$\pi_{t}\in\{1,2,\ldots,K\}$ denoting that at each time $t=1,2,\ldots,N$,
the arm $\pi_{t}$ is selected to pull. Note that $\pi_{t}$ depends
only on previous $t-1$ observations. The objective of a good policy
$\pi$ is to minimize the \emph{regret} 
\begin{equation}
R_{N}(\pi)=\sum_{k=1}^{K}(\mu_{*}-\mu_{k})\mathbb{E}N_{k}=\sum_{t=1}^{N}(\mu_{*}-\mu_{\pi_{t}}).
\end{equation}

Note that for a data-driven policy $\hat{\pi}$, the regret monotonically
increases with respect to $N$. Hence, minimizing the growth rate
of $R_{N}$ becomes an important criterion which will be considered
later.

\begin{algorithm} 
	\caption{Sub-sample Mean Comparisons.} 
	\label{alg:SMC} 
	\begin{algorithmic}[1]
		\Require 
		The set of configurations $\mathcal{I}=\{1,\ldots,K\}$,
parameter $c_{n}$, minimum budget $b$.
		\Ensure 
		$\hat\pi_1,\ldots,\hat\pi_N\in\mathcal{I}$.
		\State $r=1$, evaluate each configuration with budget $b$. 
		\For{$r=2,3,\ldots$} 
		\State The configuration with the most budgets
is denoted by $\zeta^{r}$ and called the leader; 
		\For{$k\neq\zeta^r$}
		\State Evaluate the $k$-th configuration
with one more budget $b$ if it is ``better'' than the $\zeta^{r}$-th
configuration.
		\EndFor
		\State If there is no configuration ``better''
than the leader, evaluate the leader with one more budget $b$.
		\EndFor 
	\end{algorithmic} 
\end{algorithm}

Then we introduce an efficient nonparametric solution to the multi-armed
bandit problem. First, we revisit the Sub-sample Mean Comparisons
(SMC) introduced in \cite{Chan2019} for the HPO case. The set of
configurations $\mathcal{I}=\{1,\ldots,K\},$minimum budget $b$ and
parameter $c_{n}$ are inputs. Output is the sequence of the configurations
$\hat{\pi}_{1},\ldots,\hat{\pi}_{N}\in\mathcal{I}$ to be evaluated
in order.

First, it is defined that the $k$-th configuration is ``better''
than the $k'$-th configuration, if one of the following conditions
holds:
\begin{enumerate}
\item {\footnotesize{}$n_{k}<n_{k'}$ and $n_{k}<c_{n}$.}{\footnotesize\par}
\item {\footnotesize{}$c_{n}\leq n_{k}<n_{k'}$ and $\bar{Y}_{1:n_{k}}^{(k)}\geq\bar{Y}_{j:(j+n_{k}-1)}^{(k')}$
, for some $1\leq j\leq n_{k'}-n_{k}+1$, where $\bar{Y}_{l:u}^{(k)}=\sum_{v=l}^{u}Y_{v}^{(k)}/(u-l+1)$.}{\footnotesize\par}
\end{enumerate}
In SMC, let $r$ denotes the round number. In the first round, all
configurations are evaluated since there is no information about them.
In round $r\geq2$, we define the leader of configurations which has
been evaluated with the most budgets. And, the $k$-th configuration
will be evaluated with one more budget $b$, if it is ``better''
than the leader. Otherwise, if there is no configuration ``better''
than the leader, the leader will be evaluated again. Hence, in each
round, there are at most $K-1$ configurations and at least one configuration
to be evaluated. Let $n^{r}$ be the total number of evaluations at
the beginning of round $r$, and $n_{k}^{r}$ be the corresponding
number from the $k$-th configuration. Then, we have  $K+r-2\leq n\leq K+(K-1)(r-2)$.

The Sub-sample Mean Comparisons(SMC) is shown in Algorithm \ref{alg:SMC}. In SMC, the parameter $c_{n}$ is a non-negative monotone increasing
threshold for SMC satisfied that $c_{n}=o(\log n)$ and $c_{n}/\log\log n\rightarrow\infty$
as $n\rightarrow\infty$. In \cite{Chan2019}, they set $c_{n}=\sqrt{\log n}$
for efficiency of SMC.

Note that when the round $r$ ends, the number of evaluations $n^{r}$
usually doesn't equal to $N$ exactly, i.e., $n^{r}<N<n^{r+1}$. For
this case, $N-n^{r}$ configurations are randomly chosen from the
$n^{r+1}-n^{r}$ configurations selected by SMC in the $r$-th round.
A main advantage of SMC is that unlike the UCB-based procedures, underlying
probability distributions need not be specified. Still, it remains
asymptotic optimal efficiency. The detailed discussion about theoretical
results refers to \cite{huang2020}.

\section{Experiments}

\subsection{Datasets, Metrics and Implementation Details}

We conduct experiments to evaluate the performance of our proposed
method AABO on three object detection datasets: COCO 2017 \cite{lin2014microsoft},
Visual Genome(VG) \cite{krishnavisualgenome}, and ADE \cite{zhou2017scene}.
COCO is a common object detection dataset with 80 object classes,
containing 118K training images (\textit{train}), 5K validation images
(\textit{val}) and 20K unannotated testing images (\textit{test-dev}).
VG and ADE are two large-scale object detection benchmarks with thousands
of object classes. For COCO, we use the \textit{train} split for training
and \textit{val} split for testing. For VG, we use release v1.4 and
synsets \cite{russakovsky2015imagenet} instead of raw names of the
categories due to inconsistent annotations. Specifically, we consider
two sets containing different target classes: VG$_{1000}$ and VG$_{3000}$,
with 1000 most frequent classes and 3000 most frequent classes respectively.
In both VG$_{1000}$ and VG$_{3000}$, we use 88K images for training,
5K images for testing, following \cite{chen2018iterative,jiang2018hybrid}.
For ADE, we consider 445 classes, use 20K images for training and
1K images for testing, following \cite{chen2018iterative,jiang2018hybrid}.
Besides, the ground-truths of ADE are given as segmentation masks,
so we first convert them to bounding boxes for all instances before
training.

As for evaluation, the results of the detection tasks are estimated
with standard COCO metrics, including mean Average Precision (mAP)
across IoU thresholds from 0.5 to 0.95 with an interval of 0.05 and
AP{\footnotesize{}$_{50}$}, AP{\footnotesize{}$_{75}$}, as well
as AP{\footnotesize{}$_{S}$}, AP{\footnotesize{}$_{M}$} and AP{\footnotesize{}$_{L}$},
which respectively concentrate on objects of small size 
($32\times 32-$), medium size ($32\times 32$ $\sim$ $96\times 96$)
and large size ($96\times 96 +$).

During anchor configurations searching, we sample anchor configurations
and train Faster-RCNN \cite{ren2015faster} (combined with FPN \cite{lin2017feature})
using the sampled anchor configurations, then compare the performance
of these models (regrading mAP as evaluation metrics) to reserve better
anchor configurations and stop the poorer ones, following the method
we proposed before. The search space is feature-map-wised as introduced.
All the experiments are conducted on 4 servers with 8 Tesla V100 GPUs,
using the Pytorch framework \cite{paszke2017automatic,mmdetection2018}.
ResNet-50 \cite{he2016deep} pre-trained on ImageNet \cite{russakovsky2015imagenet}
is used as the shared backbone networks. We use SGD (momentum = 0.9)
with batch size of 64 and train 12 epochs in total with an initial
learning rate of 0.02, and decay the learning rate by 0.1 twice during
training.

\subsection{Anchor Optimization Results}

We first evaluate the effectiveness of AABO over 3 large-scale detection
datasets: COCO \cite{lin2014microsoft}, VG \cite{krishnavisualgenome}
(including VG$_{1000}$ and VG$_{3000}$) and ADE \cite{zhou2017scene}.
We use Faster-RCNN \cite{ren2015faster} combined with FPN \cite{lin2017feature}
as our detector, and the baseline model is FPN with default anchor configurations. The results are shown in Table \ref{tab:results-3datasets}
and the optimal anchors searched out by AABO are reported in the appendix.

\begin{table}[tb]
\begin{centering}
\par\end{centering}
\caption{\label{tab:results-3datasets}The results of our proposed method on
some large-scale methods. We use Faster-RCNN \cite{ren2015faster}
combined with FPN \cite{lin2017feature} as detectors and ResNet-50
as backbones.}

\begin{centering}
\par\end{centering}
\begin{centering}
\tabcolsep 0.05in{\scriptsize{}}%
\begin{tabular}{c|c|cccccc}
\hline 
{\scriptsize{}Dataset} & {\scriptsize{}Method} & {\scriptsize{}mAP} & {\scriptsize{}AP$_{50}$} & {\scriptsize{}AP$_{75}$} & {\scriptsize{}AP$_{S}$} & {\scriptsize{}AP$_{M}$} & {\scriptsize{}AP$_{L}$}\tabularnewline
\hline 
\multirow{2}{*}{{\scriptsize{}COCO}} & {\scriptsize{}Faster-RCNN w FPN} & {\scriptsize{}36.4} & {\scriptsize{}58.2} & {\scriptsize{}39.1} & {\scriptsize{}21.3} & {\scriptsize{}40.1} & {\scriptsize{}46.5}\tabularnewline
 & {\scriptsize{}Search via AABO} & \textbf{\scriptsize{}38.8$^{+2.4}$} & \textbf{\scriptsize{}60.7$^{+2.5}$} & \textbf{\scriptsize{}41.6$^{+2.5}$} & \textbf{\scriptsize{}23.7$^{+2.4}$} & \textbf{\scriptsize{}42.5$^{+2.4}$} & \textbf{\scriptsize{}51.5$^{+5.0}$}\tabularnewline
\hline 
\multirow{2}{*}{{\scriptsize{}VG$_{1000}$}} & {\scriptsize{}Faster-RCNN w FPN} & {\scriptsize{}6.5} & {\scriptsize{}12.0} & {\scriptsize{}6.4} & {\scriptsize{}3.7} & {\scriptsize{}7.2} & {\scriptsize{}9.5}\tabularnewline
 & {\scriptsize{}Search via AABO} & \textbf{\scriptsize{}8.0}{\scriptsize{}$^{+1.5}$} & \textbf{\scriptsize{}13.2}{\scriptsize{}$^{+1.2}$} & \textbf{\scriptsize{}8.2}{\scriptsize{}$^{+1.8}$} & \textbf{\scriptsize{}4.2}{\scriptsize{}$^{+0.5}$} & \textbf{\scriptsize{}8.3}{\scriptsize{}$^{+1.1}$} & \textbf{\scriptsize{}12.0}{\scriptsize{}$^{+2.5}$}\tabularnewline
\hline 
\multirow{2}{*}{{\scriptsize{}VG$_{3000}$}} & {\scriptsize{}Faster-RCNN w FPN} & {\scriptsize{}3.7} & {\scriptsize{}6.5} & {\scriptsize{}3.6} & {\scriptsize{}2.3} & {\scriptsize{}4.9} & {\scriptsize{}6.8}\tabularnewline
 & {\scriptsize{}Search via AABO} & \textbf{\scriptsize{}4.2}{\scriptsize{}$^{+0.5}$} & \textbf{\scriptsize{}6.9}{\scriptsize{}$^{+0.4}$} & \textbf{\scriptsize{}4.6}{\scriptsize{}$^{+1.0}$} & \textbf{\scriptsize{}3.0}{\scriptsize{}$^{+0.7}$} & \textbf{\scriptsize{}5.8}{\scriptsize{}$^{+0.9}$} & \textbf{\scriptsize{}7.9}{\scriptsize{}$^{+1.1}$}\tabularnewline
\hline 
\multirow{2}{*}{{\scriptsize{}ADE}} & {\scriptsize{}Faster-RCNN w FPN} & {\scriptsize{}10.3} & {\scriptsize{}19.1} & {\scriptsize{}10.0} & {\scriptsize{}6.1} & {\scriptsize{}11.2} & {\scriptsize{}16.0}\tabularnewline
 & {\scriptsize{}Search via AABO} & \textbf{\scriptsize{}11.9}{\scriptsize{}$^{+1.6}$} & \textbf{\scriptsize{}20.7}{\scriptsize{}$^{+1.6}$} & \textbf{\scriptsize{}11.9}{\scriptsize{}$^{+1.9}$} & \textbf{\scriptsize{}7.4}{\scriptsize{}$^{+1.3}$} & \textbf{\scriptsize{}12.2}{\scriptsize{}$^{+1.0}$} & \textbf{\scriptsize{}17.5}{\scriptsize{}$^{+1.5}$}\tabularnewline
\hline 
\end{tabular}{\scriptsize\par}
\par\end{centering}
\centering{}
\end{table}

It's obvious that AABO outperforms Faster-RCNN with default anchor settings
among all the 3 datasets. Specifically, AABO improves mAP by 2.4\%
on COCO, 1.5\% on VG$_{1000}$, 0.5\% on VG$_{3000}$, and 1.6\% on
ADE. The results illustrate that the pre-defined anchor used in common-used
detectors are not optimal. Treat anchor configurations as hyper-parameters
and optimize them using AABO can assist to determine better anchor
settings and improve the performance of the detectors without increasing
the complexity of the network.

Note that the searched anchors increase all the AP metrics, and the
improvements on AP$_{L}$ are always more significant than AP$_{S}$
and AP$_{M}$ : AABO boosts AP$_{L}$ by 5\% on COCO, 2.5\% on VG$_{1000}$,
1.1\% on VG$_{3000}$, and 1.5\% on ADE. These results indicate that
anchor configurations determined by AABO concentrate better on all
objects, especially on the larger ones. It can also be found that
AABO is especially useful for the large-scale object detection dataset
such as VG$_{3000}$. We conjecture that this is because the searched
anchors can better capture the various sizes and shapes of objects
in a large number of categories.

\subsection{Benefit of the Optimal Anchor Settings on SOTA Methods}

After searching out optimal anchor configurations via AABO, we apply
them on several other backbones and detectors to study the generalization
property of the anchor settings. For backbone, we change ResNet-50
\cite{he2016deep} to ResNet-101 \cite{he2016deep} and ResNeXt-101
\cite{xie2017aggregated}, with detector (FPN) and other conditions
constant. For detectors, we apply our searched anchor settings on
several state-of-the-art detectors: a) Mask RCNN \cite{He2017Mask},
b) RetinaNet \cite{lin2017focal}, which is a one-stage detector,
c) DCNv2 \cite{zhu2018deformable}, and d) Hybrid Task Cascade (HTC)
\cite{chen2019hybrid}, with different backbones: ResNet-101 and ResNeXt-101.
All the experiments are conducted on COCO.

The results are reported in Table \ref{tab:transfer}. We can observe
that the optimal anchors can consistently boost the performance of
SOTA detectors, whether one-stage methods or two-stage methods. Concretely,
the optimal anchors bring 2.1\% mAP improvement on FPN with ResNet-101,
1.9\% on FPN with ResNeXt-101, 2.0\% on Mask RCNN, 1.4\% on RetinaNet,
2.4\% on DCNv2, and 1.4\% improvement on HTC. The results demonstrate
that our optimal anchors can be widely applicable across different
network backbones and SOTA detection algorithms, including both one-stage
and two-stage detectors. We also evaluate these optimized SOTA detectors
on COCO \textit{test-dev}, and the results are reported in the appendix.
The performance improvements on \textit{val} split and \textit{test-dev}
are consistent.

\begin{table}[tb]
\begin{centering}
\par\end{centering}
\caption{\label{tab:transfer}Improvements on SOTA detectors over COCO\textit{
val}. The optimal anchors are applied on several SOTA detectors with
different backbones.}

\begin{centering}
\par\end{centering}
\begin{centering}
\tabcolsep 0.05in{\scriptsize{}}%
\begin{tabular}{c|c|cccccc}
\hline 
\multicolumn{2}{c|}{{\scriptsize{}Model}} & {\scriptsize{}mAP} & {\scriptsize{}AP$_{50}$} & {\scriptsize{}AP$_{75}$} & {\scriptsize{}AP$_{S}$} & {\scriptsize{}AP$_{M}$} & {\scriptsize{}AP$_{L}$}\tabularnewline
\hline 
\multirow{2}{*}{{\scriptsize{}FPN\cite{lin2017feature} w r101}} & {\scriptsize{}Default} & {\scriptsize{}38.4} & {\scriptsize{}60.1} & {\scriptsize{}41.7} & {\scriptsize{}21.6} & {\scriptsize{}42.7} & {\scriptsize{}50.1}\tabularnewline
 & {\scriptsize{}Searched} & \textbf{\scriptsize{}40.5}{\scriptsize{}$^{+2.1}$} & \textbf{\scriptsize{}61.8} & \textbf{\scriptsize{}43.3} & \textbf{\scriptsize{}23.4} & \textbf{\scriptsize{}43.6} & \textbf{\scriptsize{}51.3}\tabularnewline
\hline 
\multirow{2}{*}{{\scriptsize{}FPN\cite{lin2017feature} w x101}} & {\scriptsize{}Default} & {\scriptsize{}40.1} & {\scriptsize{}62.0} & {\scriptsize{}43.8} & {\scriptsize{}24.0} & {\scriptsize{}44.8} & {\scriptsize{}51.7}\tabularnewline
 & {\scriptsize{}Searched} & \textbf{\scriptsize{}42.0}{\scriptsize{}$^{+1.9}$} & \textbf{\scriptsize{}63.9} & \textbf{\scriptsize{}65.1} & \textbf{\scriptsize{}25.2} & \textbf{\scriptsize{}46.3} & \textbf{\scriptsize{}54.4}\tabularnewline
\hline 
\multirow{2}{*}{{\scriptsize{}Mask RCNN\cite{He2017Mask} w r101}} & {\scriptsize{}Default} & {\scriptsize{}40.3} & {\scriptsize{}61.5} & {\scriptsize{}44.1} & {\scriptsize{}22.2} & {\scriptsize{}44.8} & {\scriptsize{}52.9}\tabularnewline
 & {\scriptsize{}Searched} & \textbf{\scriptsize{}42.3}{\scriptsize{}$^{+2.0}$} & \textbf{\scriptsize{}63.6} & \textbf{\scriptsize{}46.3} & \textbf{\scriptsize{}26.1} & \textbf{\scriptsize{}46.3} & \textbf{\scriptsize{}55.3}\tabularnewline
\hline 
\multirow{2}{*}{{\scriptsize{}RetinaNet\cite{lin2017focal} w r101}} & {\scriptsize{}Default} & {\scriptsize{}38.1} & {\scriptsize{}58.1} & {\scriptsize{}40.6} & {\scriptsize{}20.2} & {\scriptsize{}41.8} & {\scriptsize{}50.8}\tabularnewline
 & {\scriptsize{}Searched} & \textbf{\scriptsize{}39.5}{\scriptsize{}$^{+1.4}$} & \textbf{\scriptsize{}60.2} & \textbf{\scriptsize{}41.9} & \textbf{\scriptsize{}21.7} & \textbf{\scriptsize{}42.7} & \textbf{\scriptsize{}53.7}\tabularnewline
\hline 
\multirow{2}{*}{{\scriptsize{}DCNv2\cite{zhu2018deformable} w x101}} & {\scriptsize{}Default} & {\scriptsize{}43.4} & {\scriptsize{}61.3} & {\scriptsize{}47.0} & {\scriptsize{}24.3} & {\scriptsize{}46.7} & {\scriptsize{}58.0}\tabularnewline
 & {\scriptsize{}Searched} & \textbf{\scriptsize{}45.8}{\scriptsize{}$^{+2.4}$} & \textbf{\scriptsize{}67.5} & \textbf{\scriptsize{}49.7} & \textbf{\scriptsize{}28.9} & \textbf{\scriptsize{}49.4} & \textbf{\scriptsize{}60.9}\tabularnewline
\hline 
\multirow{2}{*}{{\scriptsize{}HTC\cite{chen2019hybrid} w x101}} & {\scriptsize{}Default} & {\scriptsize{}46.8} & {\scriptsize{}66.2} & {\scriptsize{}51.2} & {\scriptsize{}28.0} & {\scriptsize{}50.6} & {\scriptsize{}62.0}\tabularnewline
 & {\scriptsize{}Searched} & \textbf{\scriptsize{}48.2}{\scriptsize{}$^{+1.4}$} & \textbf{\scriptsize{}67.3} & \textbf{\scriptsize{}52.2} & \textbf{\scriptsize{}28.6} & \textbf{\scriptsize{}51.9} & \textbf{\scriptsize{}62.7}\tabularnewline
\hline 
\end{tabular}{\scriptsize\par}
\par\end{centering}
\centering{}
\end{table}

\subsection{Comparison with Other Optimization Methods}

\textbf{Comparison with other anchor initialization methods. }In this
section, we compare AABO with several existing anchor initialization
methods: a) Pre-define anchor settings, which is used in most modern
detectors. b) Use k-means to obtain clusters and treat them as default
anchors, which is used in YOLOv2 \cite{redmon2017yolo9000}. c) Use
random search to determine anchors. d) Use AABO combined with Hyperband
\cite{Li2016Hyperband} to determine anchors. e) Use AABO (combined
with sub-sampling) to determine anchors. Among all these methods,
the latter three use HPO methods to select anchor boxes automatically,
while a) and b) use naive methods like handcrafting and k-means. The
results are recorded in Table \ref{tab:results-3methods}.

Among all these anchor initialization methods, our proposed approach
can boost the performance most significantly, bring 2.4\% improvement
than using default anchors, while the improvements of other methods
including statistical methods and previous HPO methods are less remarkable.
The results illustrate that the widely used anchor initialization
approaches might be sub-optimal, while AABO can fully utilize the
ability of advanced detection systems.

\begin{table}[tb]
\begin{centering}
\par\end{centering}
\caption{\label{tab:results-3methods}Comparison with other anchor initialization
methods. Here HB denotes Hyperband while SS denotes sub-sampling.
The experiments are conducted on COCO, using FPN (with ResNet-50) as detector. Note that feature-map-wised search space is much huger
than the single one, so random search fails to converge.}

\begin{centering}
\par\end{centering}
\begin{centering}
\tabcolsep 0.05in{\footnotesize{}}%
\begin{tabular}{c|c|cccccc}
\hline 
\multicolumn{2}{c|}{{\scriptsize{}Methods to Determine Anchor}} & {\scriptsize{}mAP} & {\scriptsize{}AP$_{50}$} & {\scriptsize{}AP$_{75}$} & {\scriptsize{}AP$_{S}$} & {\scriptsize{}AP$_{M}$} & {\scriptsize{}AP$_{L}$}\tabularnewline
\hline 
{\scriptsize{}Manual Methods} & {\scriptsize{}Pre-defined} & {\scriptsize{}36.4} & {\scriptsize{}58.2} & {\scriptsize{}39.1} & {\scriptsize{}21.3} & {\scriptsize{}40.1} & {\scriptsize{}46.5}\tabularnewline
\hline 
{\scriptsize{}Statistical Methods} & {\scriptsize{}K-Means} & {\scriptsize{}37.0$^{+0.6}$} & {\scriptsize{}58.9} & {\scriptsize{}39.8} & {\scriptsize{}21.9} & {\scriptsize{}40.5} & {\scriptsize{}48.5}\tabularnewline
\hline 
\multirow{3}{*}{{\scriptsize{}HPO}} & {\scriptsize{}Random Search} & {\scriptsize{}11.5$^{-24.9}$} & {\scriptsize{}19.1} & {\scriptsize{}8.2} & {\scriptsize{}4.5} & {\scriptsize{}10.1} & {\scriptsize{}13.6}\tabularnewline
 & {\scriptsize{}AABO w HB} & {\scriptsize{}38.2$^{+1.8}$} & {\scriptsize{}59.3} & {\scriptsize{}40.7} & {\scriptsize{}22.6} & {\scriptsize{}42.1} & {\scriptsize{}50.1}\tabularnewline
 & {\scriptsize{}AABO w SS} & \textbf{\scriptsize{}38.8}{\scriptsize{}$^{+2.4}$} & \textbf{\scriptsize{}60.7} & \textbf{\scriptsize{}41.6} & \textbf{\scriptsize{}23.7} & \textbf{\scriptsize{}42.5} & \textbf{\scriptsize{}51.5}\tabularnewline
\hline 
\end{tabular}{\footnotesize\par}
\par\end{centering}
\centering{}
\end{table}

\textbf{Comparison with other HPO methods}. As Table \ref{tab:Computation}
shows, our proposed method could find better anchor configurations
in fewer trials and can improve the performance of the detector significantly:
With single search space, AABO combined with HB and SS boosts the
mAP of FPN \cite{lin2017feature} from 36.4\% to 37.8\% and 38.3\%
respectively, while random search only boosts 36.4\% to 37.2\%. With
feature-map-wised search space, AABO combined with HB and SS can obtain
38.2\% and 38.8\% mAP respectively, while random search fails to converge
due to the huge and flexible search space. The results illustrate
the effectiveness and the high efficiency of our proposed approach.

\begin{table}[tb]
\begin{centering}
\par\end{centering}
\begin{centering}
{\footnotesize{}\caption{\label{tab:Computation}The search efficiency of some HPO methods on COCO with FPN (with ResNet-50). HB denotes Hyperband while SS denotes sub-sampling.}
}{\footnotesize\par}
\par\end{centering}
\begin{centering}
{\footnotesize{}\tabcolsep 0.005in}{\scriptsize{}}%
\begin{tabular}{c|c|c|c}
\hline 
\multirow{2}{*}{{\scriptsize{}Search Space}} & {\scriptsize{}Search} & {\scriptsize{}mAP of} & {\scriptsize{}number of searched}\tabularnewline
 & {\scriptsize{}Method} & {\scriptsize{}optimal anchor} & {\scriptsize{}parameters}\tabularnewline
\hline 
\multirow{3}{*}{{\scriptsize{}Single}} & {\scriptsize{}Random} & {\scriptsize{}37.2} & {\scriptsize{}100}\tabularnewline
 & {\scriptsize{}AABO w HB} & {\scriptsize{}37.8$^{+0.6}$} & {\scriptsize{}64}\tabularnewline
 & {\scriptsize{}AABO w SS} & {\scriptsize{}38.3$^{+1.1}$} & {\scriptsize{}64}\tabularnewline
\hline 
\multirow{3}{*}{{\scriptsize{}Feature-Map-Wised}} & {\scriptsize{}Random} & {\scriptsize{}11.5} & {\scriptsize{}100}\tabularnewline
 & {\scriptsize{}AABO w HB} & {\scriptsize{}38.2$^{+26.7}$} & {\scriptsize{}64}\tabularnewline
 & {\scriptsize{}AABO w SS} & \textbf{\scriptsize{}38.8}{\scriptsize{}$^{+27.3}$} & {\scriptsize{}64}\tabularnewline
\hline 
\end{tabular}{\scriptsize\par}
\par\end{centering}
\centering{}
\end{table}

\textbf{Comparison with other anchor optimization methods.} We also
compare AABO with some previous anchor optimization methods like \cite{Zhong2018}
and MetaAnchor \cite{Tong2018MetaAnchor}. As shown in Table \ref{tab:results-3methods-1},
all these methods can boost the performance of detectors, while our
method brings 2.4\% mAP improvement on Faster-RCNN and 1.4\% on RetinaNet,
and the other two methods only bring 1.0\% improvement, which demonstrates
the superiority of our proposed AABO.

\begin{table}[tb]
\begin{centering}
\par\end{centering}
\caption{\label{tab:results-3methods-1}Comparison with previous anchor optimization
methods: Zhong's method \cite{Zhong2018} and MetaAnchor \cite{Tong2018MetaAnchor}.
The results are extracted from their papers respectively. Note that
we search optimal anchors for Faster-RCNN originally, then directly
apply them to RetinaNet. Therefore, the performance improvement on
RetinaNet is not as significant as Faster-RCNN, but still better than
the other methods.}

\begin{centering}
\par\end{centering}
\begin{centering}
\tabcolsep 0.05in{\scriptsize{}}%
\begin{tabular}{c|cccc}
\hline 
\multirow{2}{*}{{\scriptsize{}Methods}} & {\scriptsize{}YOLOv2} & {\scriptsize{}RetinaNet} & {\scriptsize{}Faster-RCNN} & {\scriptsize{}RetinaNet}\tabularnewline
 & {\scriptsize{}w Zhong's Method \cite{Zhong2018}} & {\scriptsize{}w MetaAnchor \cite{Tong2018MetaAnchor}} & {\scriptsize{}w AABO} & {\scriptsize{}w AABO}\tabularnewline
\hline 
{\scriptsize{}mAP of Baseline} & {\scriptsize{}23.5} & {\scriptsize{}36.9} & {\scriptsize{}36.4} & {\scriptsize{}38.1}\tabularnewline
\hline 
{\scriptsize{}mAP after Optimization} & {\scriptsize{}24.5$^{+1.0}$} & {\scriptsize{}37.9$^{+1.0}$} & {\scriptsize{}38.8$^{+2.4}$} & \textbf{\scriptsize{}39.5$^{+1.4}$}\tabularnewline
\hline 
\end{tabular}{\scriptsize\par}
\par\end{centering}
\centering{}
\end{table}

\subsection{Ablation Study}

In this section, we study the effects of all the components used in
AABO: a) Treat anchor settings as hyper-parameters, then use HPO methods
to search them automatically. b) Use Bayesian Optimization method.
c) Use sub-sampling method to determine the reserved anchors. d) Feature-map-wised
search space.

As shown in Table \ref{tab:abalation}, using HPO methods such as
random search to optimize anchors can bring 0.8\% performance
improvement, which indicates the default anchors are sub-optimal.
Using single search space (not feature-map-wised), AABO combined with
HB brings 1.4\% mAP improvement, and AABO combined with SS
brings 1.9\% mAP improvement, which demonstrates the advantage of
BO and accurate estimation of acquisition function. Besides, our tight
and adaptive feature-map-wised search space can give a guarantee to
search out better anchors with limited computation resources, and
brings about 0.5\% mAP improvement as well. Our method AABO can increase
mAP by 2.4\% overall.

\begin{table}[tb]
\begin{centering}
\par\end{centering}
\caption{\label{tab:abalation}Regard Bayesian Optimization (BO), Sub-sampling
(SS), Feature-map-wised search space as key components of AABO, we
study the effectiveness of all these components. HB denotes Hyperband
and SS denotes sub-sampling. The experiments are conducted on COCO,
using FPN (with ResNet-50) as detector. Note that random search
fails to converge due to the huge feature-map-wised search
space.}

\begin{centering}
\par\end{centering}
\begin{centering}
\tabcolsep 0.03in{\scriptsize{}}%
\begin{tabular}{c|cccc|cccccc}
\hline 
{\scriptsize{}Model} & {\scriptsize{}Search} & {\scriptsize{}BO} & {\scriptsize{}SS} & {\scriptsize{}Feature-map-wised} & {\scriptsize{}mAP} & {\scriptsize{}AP$_{50}$} & {\scriptsize{}AP$_{75}$} & {\scriptsize{}AP$_{S}$} & {\scriptsize{}AP$_{M}$} & {\scriptsize{}AP$_{L}$}\tabularnewline
\hline 
{\scriptsize{}Default} &  &  &  &  & {\scriptsize{}36.4} & {\scriptsize{}58.2} & {\scriptsize{}39.1} & {\scriptsize{}21.3} & {\scriptsize{}40.1} & {\scriptsize{}46.5}\tabularnewline
\hline 
\multirow{2}{*}{{\scriptsize{}Random Search}} & \textbf{\scriptsize{}$\checked$} &  &  &  & {\scriptsize{}37.2$^{+0.8}$} & {\scriptsize{}58.8} & {\scriptsize{}39.9} & {\scriptsize{}21.7} & {\scriptsize{}40.6} & {\scriptsize{}48.1}\tabularnewline
 & \textbf{\scriptsize{}$\checked$} &  &  & \textbf{\scriptsize{}$\checked$} & {\scriptsize{}11.5$^{-24.9}$} & {\scriptsize{}19.1} & {\scriptsize{}8.2} & {\scriptsize{}4.5} & {\scriptsize{}10.1} & {\scriptsize{}13.6}\tabularnewline
\hline 
\multirow{2}{*}{{\scriptsize{}AABO w HB}} & \textbf{\scriptsize{}$\checked$} & \textbf{\scriptsize{}$\checked$} &  &  & {\scriptsize{}37.8$^{+1.4}$} & {\scriptsize{}58.9} & {\scriptsize{}40.4} & {\scriptsize{}22.7} & {\scriptsize{}41.3} & {\scriptsize{}49.9}\tabularnewline
 & \textbf{\scriptsize{}$\checked$} & \textbf{\scriptsize{}$\checked$} &  & \textbf{\scriptsize{}$\checked$} & {\scriptsize{}38.2$^{+1.8}$} & {\scriptsize{}59.3} & {\scriptsize{}40.7} & {\scriptsize{}22.6} & {\scriptsize{}42.1} & {\scriptsize{}50.1}\tabularnewline
\hline 
\multirow{2}{*}{{\scriptsize{}AABO w SS}} & \textbf{\scriptsize{}$\checked$} & \textbf{\scriptsize{}$\checked$} & \textbf{\scriptsize{}$\checked$} &  & {\scriptsize{}38.3$^{+1.9}$} & {\scriptsize{}59.6} & {\scriptsize{}40.9} & {\scriptsize{}22.9} & {\scriptsize{}42.2} & {\scriptsize{}50.8}\tabularnewline
 & \textbf{\scriptsize{}$\checked$} & \textbf{\scriptsize{}$\checked$} & \textbf{\scriptsize{}$\checked$} & \textbf{\scriptsize{}$\checked$} & \textbf{\scriptsize{}38.8}{\scriptsize{}$^{+2.4}$} & \textbf{\scriptsize{}60.7} & \textbf{\scriptsize{}41.6} & \textbf{\scriptsize{}23.7} & \textbf{\scriptsize{}42.5} & \textbf{\scriptsize{}51.5}\tabularnewline
\hline 
\end{tabular}{\scriptsize\par}
\par\end{centering}
\centering{}
\end{table}

\section{Conclusion}

In this work, we propose AABO, an adaptive anchor box optimization
method for object detection via Bayesian sub-sampling, where optimal
anchor configurations for a certain dataset and detector are determined
automatically without manually adjustment. We demonstrate that AABO
outperforms both hand-adjusted methods and HPO methods on popular
SOTA detectors over multiple datasets, which indicates that anchor
configurations play an important role in object detection frameworks
and our proposed method could help exploit the potential of detectors
in a more effective way.

\bibliographystyle{splncs04}
\bibliography{egbib}

\begin{thebibliography}{10}
\providecommand{\url}[1]{\texttt{#1}}
\providecommand{\urlprefix}{URL }
\providecommand{\doi}[1]{https://doi.org/#1}

\bibitem{Bergstra2011Algorithms}
Bergstra, J.S., Bardenet, R., Bengio, Y., K{\'e}gl, B.: Algorithms for
  hyper-parameter optimization. In: NIPS (2011)

\bibitem{bhagavatula2017faster}
Bhagavatula, C., Zhu, C., Luu, K., Savvides, M.: Faster than real-time facial
  alignment: A 3d spatial transformer network approach in unconstrained poses.
  In: ICCV (2017)

\bibitem{Cai2019Cascade}
Cai, Z., Vasconcelos, N.: Cascade r-cnn: High quality object detection and
  instance segmentation. IEEE Transactions on Pattern Analysis and Machine
  Intelligence  (2019)

\bibitem{chabot2017deep}
Chabot, F., Chaouch, M., Rabarisoa, J., Teuliere, C., Chateau, T.: Deep manta:
  A coarse-to-fine many-task network for joint 2d and 3d vehicle analysis from
  monocular image. In: CVPR (2017)

\bibitem{Chan2019}
Chan, H.P.: The multi-armed bandit problem: An efficient non-parametric
  solution. Annals of Statistics p. To appear (2019)

\bibitem{mmdetection2018}
Chen, K., Pang, J., Wang, J., Xiong, Y., Li, X., Sun, S., Feng, W., Liu, Z.,
  Shi, J., Ouyang, W., Loy, C.C., Lin, D.: mmdetection.
  \url{https://github.com/open-mmlab/mmdetection} (2018)

\bibitem{chen2019hybrid}
Chen, K., Pang, J., Wang, J., Xiong, Y., Li, X., Sun, S., Feng, W., Liu, Z.,
  Shi, J., Ouyang, W., Loy, C.C., Lin, D.: Hybrid task cascade for instance
  segmentation. In: IEEE Conference on Computer Vision and Pattern Recognition
  (2019)

\bibitem{chen2018iterative}
Chen, X., Li, L.J., Fei-Fei, L., Gupta, A.: Iterative visual reasoning beyond
  convolutions. In: CVPR (2018)

\bibitem{dai2016r}
Dai, J., Li, Y., He, K., Sun, J.: R-fcn: Object detection via region-based
  fully convolutional networks. In: NIPS (2016)

\bibitem{Falkner2018}
Falkner, S., Klein, A., Hutter, F.: Bohb: Robust and efficient hyperparameter
  optimization at scale. arXiv preprint arXiv:1807.01774  (2018)

\bibitem{He2017Mask}
He, K., Gkioxari, G., Dollar, P., Girshick, R.: Mask r-cnn. In: 2017 IEEE
  International Conference on Computer Vision (ICCV) (2017)

\bibitem{he2016deep}
He, K., Zhang, X., Ren, S., Sun, J.: Deep residual learning for image
  recognition. In: CVPR (2016)

\bibitem{huang2020}
{Huang}, Y., {Li}, Y., {Li}, Z., {Zhang}, Z.: {An Asymptotically Optimal
  Multi-Armed Bandit Algorithm and Hyperparameter Optimization}. arXiv e-prints
  arXiv:2007.05670 (2020)

\bibitem{jamieson2016non}
Jamieson, K., Talwalkar, A.: Non-stochastic best arm identification and
  hyperparameter optimization. In: Artificial Intelligence and Statistics. pp.
  240--248 (2016)

\bibitem{jiang2018hybrid}
Jiang, C., Xu, H., Liang, X., Lin, L.: Hybrid knowledge routed modules for
  large-scale object detection. In: NIPS (2018)

\bibitem{krishnavisualgenome}
Krishna, R., Zhu, Y., Groth, O., Johnson, J., Hata, K., Kravitz, J., Chen, S.,
  Kalantidis, Y., Li, L.J., Shamma, D.A., Bernstein, M., Fei-Fei, L.: Visual
  genome: Connecting language and vision using crowdsourced dense image
  annotations. International Journal of Computer Vision  (2016)

\bibitem{Li2016Hyperband}
Li, L., Jamieson, K., Desalvo, G., Rostamizadeh, A., Talwalkar, A.: Hyperband:
  A novel bandit-based approach to hyperparameter optimization. Journal of
  Machine Learning Research  \textbf{18},  1--52 (2016)

\bibitem{lin2017feature}
Lin, T.Y., Doll{\'a}r, P., Girshick, R., He, K., Hariharan, B., Belongie, S.:
  Feature pyramid networks for object detection. In: CVPR (2017)

\bibitem{lin2017focal}
Lin, T.Y., Goyal, P., Girshick, R., He, K., Doll{\'a}r, P.: Focal loss for
  dense object detection. In: ICCV. pp. 2980--2988 (2017)

\bibitem{lin2014microsoft}
Lin, T.Y., Maire, M., Belongie, S., Hays, J., Perona, P., Ramanan, D.,
  Doll{\'a}r, P., Zitnick, C.L.: Microsoft coco: Common objects in context. In:
  ECCV (2014)

\bibitem{liu2016ssd}
Liu, W., Anguelov, D., Erhan, D., Szegedy, C., Reed, S., Fu, C.Y., Berg, A.C.:
  Ssd: Single shot multibox detector. In: ECCV (2016)

\bibitem{luo2014switchable}
Luo, P., Tian, Y., Wang, X., Tang, X.: Switchable deep network for pedestrian
  detection. In: CVPR (2014)

\bibitem{mendoza2016towards}
Mendoza, H., Klein, A., Feurer, M., Springenberg, J.T., Hutter, F.: Towards
  automatically-tuned neural networks. In: Workshop on Automatic Machine
  Learning. pp. 58--65 (2016)

\bibitem{paszke2017automatic}
Paszke, A., Gross, S., Chintala, S., Chanan, G., Yang, E., DeVito, Z., Lin, Z.,
  Desmaison, A., Antiga, L., Lerer, A.: Automatic differentiation in pytorch.
  In: NIPS Workshop (2017)

\bibitem{perchet2013the}
Perchet, V., Rigollet, P.: The multi-armed bandit problem with covariates.
  Annals of Statistics  \textbf{41}(2),  693--721 (2013)

\bibitem{redmon2016you}
Redmon, J., Divvala, S., Girshick, R., Farhadi, A.: You only look once:
  Unified, real-time object detection. In: CVPR (2016)

\bibitem{redmon2017yolo9000}
Redmon, J., Farhadi, A.: Yolo9000: better, faster, stronger. In: CVPR (2017)

\bibitem{ren2015faster}
Ren, S., He, K., Girshick, R., Sun, J.: Faster r-cnn: Towards real-time object
  detection with region proposal networks. In: NIPS (2015)

\bibitem{russakovsky2015imagenet}
Russakovsky, O., Deng, J., Su, H., Krause, J., Satheesh, S., Ma, S., Huang, Z.,
  Karpathy, A., Khosla, A., Bernstein, M., et~al.: Imagenet large scale visual
  recognition challenge. IJCV  \textbf{115}(3),  211--252 (2015)

\bibitem{Snoek2012Practical}
Snoek, J., Larochelle, H., Adams, R.P.: Practical bayesian optimization of
  machine learning algorithms. In: NIPS (2012)

\bibitem{sutton2018reinforcement}
Sutton, R.S., Barto, A.G.: Reinforcement learning: An introduction. MIT press
  (2018)

\bibitem{Tong2018MetaAnchor}
Tong, Y., Zhang, X., Zhang, W., Jian, S.: Metaanchor: Learning to detect
  objects with customized anchors  (2018)

\bibitem{xie2017aggregated}
Xie, S., Girshick, R., Doll{\'a}r, P., Tu, Z., He, K.: Aggregated residual
  transformations for deep neural networks. In: CVPR. pp. 1492--1500 (2017)

\bibitem{Zhong2018}
Zhong, Y., Wang, J., Peng, J., Zhang, L.: Anchor box optimization for object
  detection. arXiv preprint arXiv:1812.00469  (2018)

\bibitem{zhou2017scene}
Zhou, B., Zhao, H., Puig, X., Fidler, S., Barriuso, A., Torralba, A.: Scene
  parsing through ade20k dataset. In: CVPR (2017)

\bibitem{zhu2018deformable}
Zhu, X., Hu, H., Lin, S., Dai, J.: Deformable convnets v2: More deformable,
  better results. arXiv preprint arXiv:1811.11168  (2018)

\end{thebibliography}

\newpage
\section*{Appendix}
\vspace{3.5mm}

\subsection*{Optimal FPN Hyper-parameters in Preliminary Analysis}

In preliminary analysis, we search for better RPN head architecture
in FPN \cite{lin2017feature} as well as anchor settings simultaneously, to compare the respective
contributions of changing RPN head architecture and anchor settings.
The optimal hyper-parameters are shown in Table \ref{tab:search-arch-anchor}.
Since the search space for RPN head architecture and anchor settings
are both relatively small, the performance increase is not very significant.

\begin{table}
{\scriptsize{}\caption{\label{tab:search-arch-anchor} The optimal hyper-parameters of FPN \cite{lin2017feature}
on COCO determined by BOHB \cite{Falkner2018} in preliminary analysis. Both RPN head
architecture and anchor boxes are different from default settings
in FPN. Note that the anchor scales here are
the basic anchor scales without multiplying the strides of the feature
maps.}
}{\scriptsize\par}
\centering{}\tabcolsep 0.06in{\scriptsize{}}%
\begin{tabular}{c|ccc}
\hline 
{\scriptsize{}Hyper-parameter} & {\scriptsize{}Conv Layers} & {\scriptsize{}Kernel Size} & {\scriptsize{}Dilation Size}\tabularnewline
{\scriptsize{}Optimal Configuration} & {\scriptsize{}2} & {\scriptsize{}5x5} & {\scriptsize{}1}\tabularnewline
\hline 
{\scriptsize{}Hyper-parameter} & {\scriptsize{}Location of ReLU} & {\scriptsize{}Anchor Scales} & {\scriptsize{}Anchor Ratios}\tabularnewline
{\scriptsize{}Optimal Configuration} & {\scriptsize{}Behind 5x5 Conv} & {\scriptsize{}2.5, 5.7, 13.4} & {\scriptsize{}2:5, 1:2, 1, 2:1, 5:2}\tabularnewline
\hline 
\end{tabular}{\scriptsize\par}
\end{table}

\subsection*{Optimal Anchor Configurations}

In this section, we record the best configurations searched out by
AABO and analyze the difference between the results of default Faster-RCNN
\cite{ren2015faster} and optimized Faster-RCNN.

As we design an adaptive feature-map-wised search space for anchor
optimization, anchor configurations distribute variously in different
layers of FPN \cite{lin2017feature}. Table \ref{tab:anchor-result}
shows the optimal anchor configurations in FPN for COCO \cite{lin2014microsoft}
dataset. We can observe that anchor scales and anchor ratios are larger
and more diverse in shallower layers of FPN, while anchors tend to
be smaller and more square in deeper layers.

\begin{table}
{\scriptsize{}\caption{\label{tab:anchor-result} The optimal anchor configurations of FPN
\cite{lin2017feature} for COCO \cite{lin2014microsoft} searched
out by AABO. There are different anchor boxes in different layers
of FPN.}
}{\scriptsize\par}
\centering{}\tabcolsep 0.01in{\scriptsize{}}%
\begin{tabular}{c|c|c|c}
\hline 
{\tiny{}FPN-Layer} & {\tiny{}Anchor Number} & {\tiny{}Anchor Scales} & {\tiny{}Anchor Ratios}\tabularnewline
\hline 
{\tiny{}Layer-1} & {\tiny{}9} & {\tiny{}\{5.2, 6.1, 3.4, 4.9, 5.8, 4.8, 14.6, 7.4, 10.3\}} & {\tiny{}\{6.0, 0.3, 0.5, 1.6, 1.7, 2.6, 0.5, 0.5, 0.6\}}\tabularnewline
{\tiny{}Layer-2} & {\tiny{}6} & {\tiny{}\{11.0, 7.6, 11.8, 4.7, 5.7, 3.8\}} & {\tiny{}\{0.2, 2.0, 2.3, 2.8, 1.0, 0.5\}}\tabularnewline
{\tiny{}Layer-3} & {\tiny{}7} & {\tiny{}\{10.5, 11.1, 8.5, 6.7, 12.4, 4.6, 3.5\}} & {\tiny{}\{0.4, 4.2, 2.2, 1.6, 2.3, 0.5, 1.1\}}\tabularnewline
{\tiny{}Layer-4} & {\tiny{}6} & {\tiny{}\{5.7, 8.5, 7.0, 11.2, 15.2, 15.5\}} & {\tiny{}\{0.3, 0.4, 0.8, 0.7, 2.9, 2.8\}}\tabularnewline
{\tiny{}Layer-5} & {\tiny{}5} & {\tiny{}\{5.0, 4.2, 14.0, 10.1, 7.8\}} & {\tiny{}\{1.1, 1.4, 0.8, 0.7, 2.5\}}\tabularnewline
\hline 
\end{tabular}{\scriptsize\par}
\end{table}

\subsection*{Improvements on SOTA Detectors over COCO Test-Dev}

After searching out optimal anchor configurations via AABO, we apply
them on several SOTA detectors to study the generalization property
of the anchor settings. In this section, we report the performance
of these optimized detectors on COCO \textit{test-dev} split. The
results are shown in Table \ref{tab:sota-test-dev}.

It can be seen that the optimal anchors
can consistently boost the performance of SOTA detectors on both COCO \textit{val} split and COCO \textit{test-dev}. Actually, the mAP of the optimized detectors on \textit{test-dev} is even higher than the mAP on
\textit{val}, which illustrates that the optimized anchor settings
could bring consistent performance improvements on \textit{val} split
and \textit{test-dev}.

\begin{table}
{\scriptsize{}\caption{\label{tab:sota-test-dev} Benefit of the optimal anchor settings
on some SOTA methods evaluated on both COCO \textit{val} and \textit{test-dev}. Here HTC* means 2x training of HTC.
The results indicate that the optimal anchors can consistently boost the performance of SOTA
detectors, whether on COCO \textit{val} or \textit{test-dev}. }
}{\scriptsize\par}
\begin{centering}
\par\end{centering}
\centering{}{\scriptsize{}}%
\setlength{\tabcolsep}{3mm}
\begin{tabular}{c|c|c|c}
\hline 
\multirow{1}{*}{{\scriptsize{}Model}} & {\scriptsize{}Anchor Setting} & {\scriptsize{}Eval on} & {\scriptsize{}mAP}\tabularnewline
\hline 
\multirow{3}{*}{{\scriptsize{}Mask RCNN\cite{He2017Mask} w r101}} & {\scriptsize{}Default} & \textit{\scriptsize{}val} & {\scriptsize{}40.3}\tabularnewline
 & {\scriptsize{}Searched via AABO} & \textit{\scriptsize{}val} & {\scriptsize{}42.3$^{+2.0}$}\tabularnewline
 & {\scriptsize{}Searched via AABO} & \textit{\scriptsize{}test-dev} & \textbf{\scriptsize{}42.6$^{+2.3}$}\tabularnewline
\hline 
\multirow{3}{*}{{\scriptsize{}DCNv2\cite{zhu2018deformable} w x101}} & {\scriptsize{}Default} & \textit{\scriptsize{}val} & {\scriptsize{}43.4}\tabularnewline
 & {\scriptsize{}Searched via AABO} & \textit{\scriptsize{}val} & {\scriptsize{}45.8$^{+2.4}$}\tabularnewline
 & {\scriptsize{}Searched via AABO} & \textit{\scriptsize{}test-dev} & \textbf{\scriptsize{}46.1$^{+2.7}$}\tabularnewline
\hline 
\multirow{3}{*}{{\scriptsize{}Cascade Mask RCNN\cite{Cai2019Cascade} w x101}} & {\scriptsize{}Default} & \textit{\scriptsize{}val} & {\scriptsize{}44.3}\tabularnewline
 & {\scriptsize{}Searched via AABO} & \textit{\scriptsize{}val} & {\scriptsize{}46.8$^{+2.5}$}\tabularnewline
 & {\scriptsize{}Searched via AABO} & \textit{\scriptsize{}test-dev} & \textbf{\scriptsize{}47.2$^{+2.9}$}\tabularnewline
\hline 
\multirow{3}{*}{{\scriptsize{}HTC{*}\cite{chen2019hybrid} w x101}} & {\scriptsize{}Default} & \textit{\scriptsize{}val} & {\scriptsize{}47.5}\tabularnewline
 & {\scriptsize{}Searched via AABO} & \textit{\scriptsize{}val} & {\scriptsize{}50.1$^{+2.6}$}\tabularnewline
 & {\scriptsize{}Searched via AABO} & \textit{\scriptsize{}test-dev} & \textbf{\scriptsize{}50.6$^{+3.1}$}\tabularnewline
\hline 
\end{tabular}{\scriptsize\par}
\end{table}

\subsection*{Additional Qualitative Results}

\begin{figure}
\begin{centering}
\includegraphics[scale=0.4]{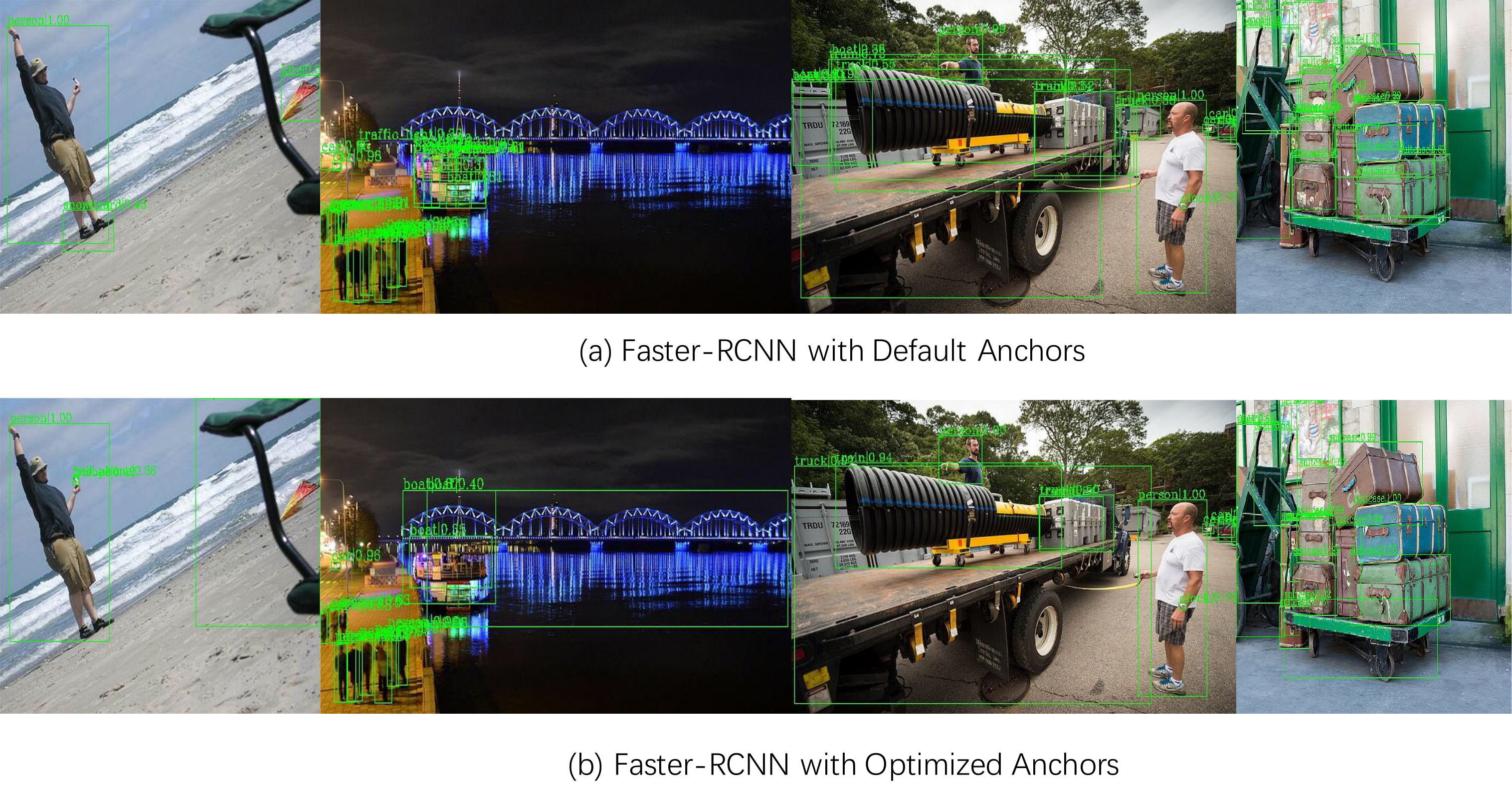}
\par\end{centering}
\caption{\label{fig:large}Some qualitative result comparison on COCO \cite{lin2014microsoft}
dataset. Using optimized anchor configurations, more large and small
objects are detected. We use ResNet-50 \cite{he2016deep} as backbones.}
\end{figure}

\begin{figure}
\begin{centering}
\includegraphics[scale=0.4]{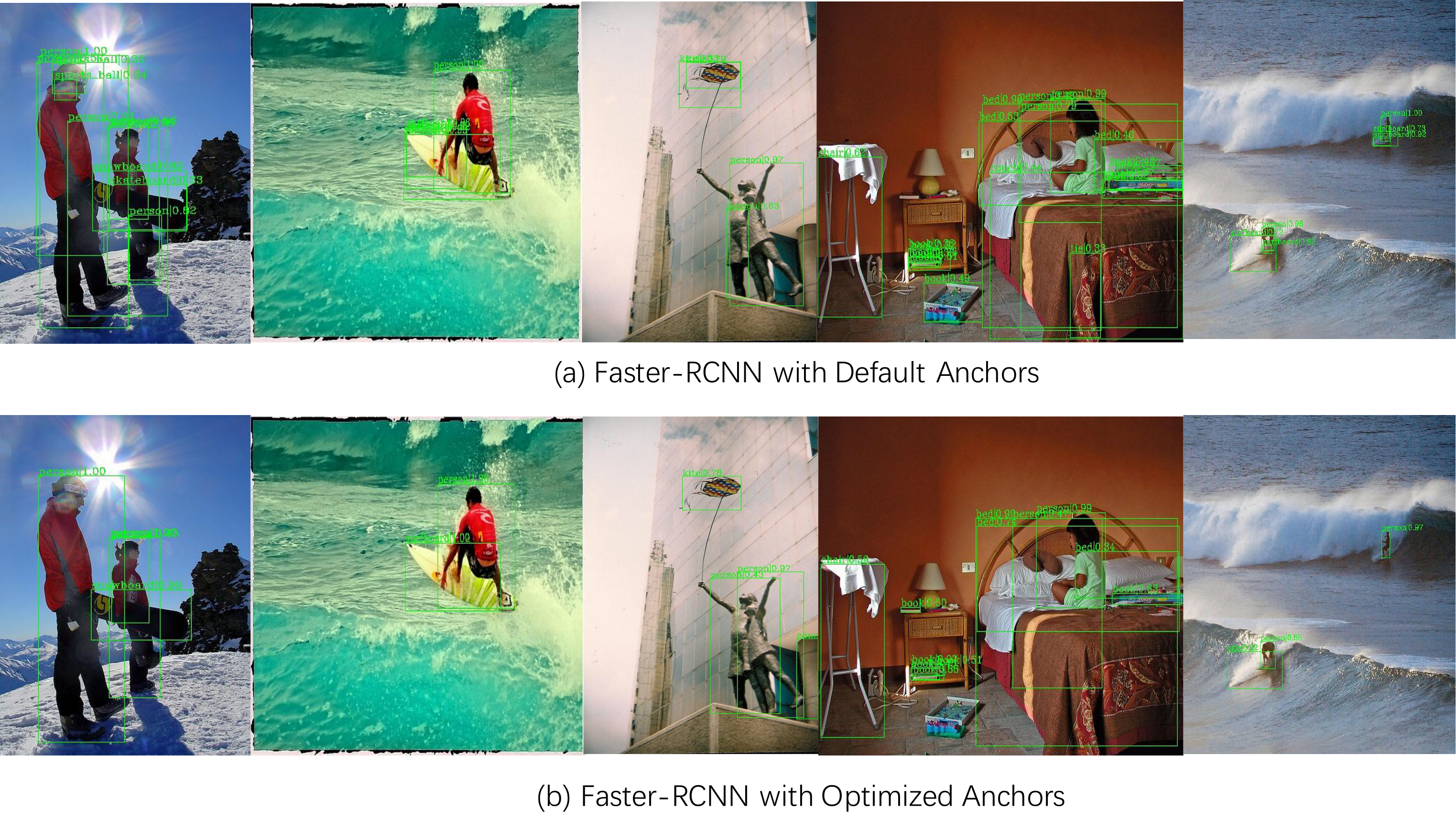}
\par\end{centering}
\caption{\label{fig:clean}Some qualitative result comparisons on COCO \cite{lin2014microsoft}
dataset. The bounding boxes given by Faster-RCNN \cite{ren2015faster}
with optimized anchor configurations are much tighter and clearer,
while bounding boxes given by default Faster-RCNN are more messy and
overlap a lot. We use ResNet-50 \cite{he2016deep} as backbones.}
\end{figure}

In this section, Figure \ref{fig:large} and Figure \ref{fig:clean} give some qualitative
result comparisons of Faster-RCNN \cite{ren2015faster} with default
anchors and optimal anchors. 

As illustrated in Figure \ref{fig:large},
more larger and smaller objects can be detected using our optimal
anchors, which demonstrates that our optimal anchors are more diverse
and suitable for a certain dataset. And there are some other differences
shown in Figure \ref{fig:clean}: Using optimal anchor settings, the
predictions of Faster-RCNN are usually tighter, more precise and concise.
While the predictions are more inaccurate and messy when using pre-defined
anchors. Specifically, there exist many bounding boxes in a certain
position, which usually are different parts of one same object and
overlap a lot.

\end{document}